\documentclass[pdflatex,sn-aps,iicol]{sn-jnl}
\usepackage{siunitx}
\usepackage{amsmath}
\usepackage{amsfonts}
\usepackage{amssymb}
\usepackage{bm}
\usepackage{scalerel}

\usepackage{graphicx}
\usepackage{standalone}
\usepackage[table]{xcolor}
\usepackage{tikz}
\usetikzlibrary{shapes, backgrounds, shapes.geometric, arrows, arrows.meta, bending, positioning}
\usepackage{circuitikz}
\usepackage{pgfplots}
\usepackage{pgfplotstable}
\usepackage{ifthen} 
\usepackage{datatool}
\usepgfplotslibrary{groupplots, statistics, fillbetween}
\pgfplotsset{compat=newest}
\usepackage[doipre={DOI:~}]{uri}

\usepackage{subcaption}
\usepackage{hyperref}
\usepackage{multirow}%
\usepackage{amsthm}%
\usepackage{mathrsfs}%
\usepackage[title]{appendix}%
\usepackage{textcomp}%
\usepackage{manyfoot}%
\usepackage{booktabs}%
\usepackage{algorithm}%
\usepackage{algorithmicx}%
\usepackage{algpseudocode}%
\usepackage{listings}%

\usepackage{lineno}

\usepackage{media9}

\usepackage{xr}
\usetikzlibrary{external}
\def\app#1#2{%
  \mathrel{%
    \setbox0=\hbox{$#1\sim$}%
    \setbox2=\hbox{%
      \rlap{\hbox{$#1\propto$}}%
      \lower1.1\ht0\box0%
    }%
    \raise0.25\ht2\box2%
  }%
}

\begin{document}

% Possible titles
\title[Article Title]{Active inference as a model of collision avoidance behavior in human drivers}

\author[1]{\fnm{Julian F.} \sur{Schumann}}

\author[2]{\fnm{Johan} \sur{Engström}}\email{jengstrom@waymo.com}

\author[2]{\fnm{Leif} \sur{Johnson}}

\author[2]{\fnm{Matthew} \sur{O'Kelly}}

\author[2]{\fnm{Joao} \sur{Messias}}

\author[1]{\fnm{Jens} \sur{Kober}}

\author[1]{\fnm{Arkady} \sur{Zgonnikov}}

\affil[1]{\orgdiv{Department of Cognitive Robotics}, \orgname{Delft University of Technology}, \country{Netherlands}}
\affil[2]{\orgname{Waymo LLC}, \city{Mountain View}, \state{CA}, \country{USA}}

% \affil[3]{\orgdiv{Department}, \orgname{Organization}, \orgaddress{\street{Street}, \city{City}, \postcode{610101}, \state{State}, \country{Country}}}


\abstract{Collision avoidance -- involving a rapid threat detection and quick execution of the appropriate evasive maneuver -- is a critical aspect of driving. However, existing models of human collision avoidance behavior are fragmented, focusing on specific scenarios or only describing certain aspects of the avoidance behavior, such as response times.
This paper addresses these gaps by proposing a computational cognitive model of human collision avoidance behavior based on active inference.
Active inference provides a unified approach to modeling human behavior: the minimization of free energy. Building on prior active inference work, our model incorporates established cognitive mechanisms such as evidence accumulation to simulate human responses in three distinct collision avoidance scenarios: front-to-rear lead vehicle braking, lateral incursion by an oncoming vehicle, and another vehicle failing to yield at an intersection. We demonstrate that our model explains a wide range of  empirical findings on human collision avoidance behavior. Specifically, the model closely reproduces both aggregate results from meta-analyses previously reported in the literature and detailed, scenario-specific effects observed in two recent driving simulator studies, including response timing, maneuver selection, and execution. Our results highlight the potential of active inference as a generalizable framework for understanding and modeling human behavior in complex real-life driving tasks.}



\maketitle

Collision avoidance is a critical skill for human drivers. It involves the rapid detection of threats (such as a vehicle ahead suddenly braking) and deciding on an appropriate evasive maneuver (for instance, braking or swerving). These maneuvers are complex, requiring not only precise execution but also continuous adjustments as the situation evolves. Furthermore, drivers need to account for the uncertainty in the future behavior of other road users: for example, will an oncoming vehicle encroaching into my lane continue across or move back to its own lane? Understanding how humans avoid collisions in traffic can provide insights into high-stakes, split-second decision making which has substantial implications for traffic safety. 

Behavior models play a key role both in understanding the mechanisms of human collision avoidance and in improving traffic safety. These models are applied in diverse contexts, such as collision risk estimation~\cite{bargman_how_2015}, understanding effects of driver distraction~\cite{engstrom_simulating_2018}, modeling take-over behavior~\cite{bianchi_piccinini_how_2020}, representing human agents in simulated test environments~\cite{montali_waymo_2023}, and providing behavioral benchmarks for automated vehicles~\cite{olleja2025validation, engstrom_modeling_2024}. Besides immediate practical applications, modeling human collision avoidance behavior is an interesting subject of study in its own right. Being a highly complex and dynamic task with extremely high stakes, collision avoidance provides a unique testbed for theories and models of cognition previously not validated in the real world~\cite{matusz_are_2019, shamay-tsoory_real-life_2019,carvalho_naturalistic_2025}.

Most existing computational models of human collision avoidance are mechanistic, that is, are based on the explicit modeling of cognitive mechanisms underlying response timing and evasive maneuvering. However, they are typically fragmented, focusing either on specific scenario types (e.g., front-to-rear conflicts~\cite{xue_using_2018,svard_computational_2021, guo_safety_2021} or merging~\cite{siebinga_model_2024-1}), specific explanatory factors (such as off-road glances~\cite{bargman_how_2015, markkula_farewell_2016} or cognitive load~\cite{engstrom_simulating_2018}), 
or only reproduce certain aspects of human behavior (such as response times~\cite{xue_using_2018,bianchi_piccinini_how_2020, svard_computational_2021,wei_modeling_2022,engstrom_modeling_2024} or the extent of steering~\cite{li_modeling_2023}). 
Altogether, these models cover a wide range of scenarios and diverse aspects of collision avoidance behavior. Yet, each one of these mechanistic models on its own is highly specific: they are not designed to generalize to multiple scenarios or describe multiple aspects of human behavior.

Recently, machine learning models based on large datasets of human driving have demonstrated the ability to generalize across a wide range of traffic scenarios~\cite{yuan_agentformer_2021,suo2021trafficsim,gu_stochastic_2022,igl2022symphony,meszaros_trajflow_2024, aydemir_adapt_2023}. Because such models typically generate full motion trajectories, they also have the potential to represent multiple aspects of collision avoidance behavior and not just a single metric of interest. However, a key challenge is that safety-critical behavior such as collision avoidance is often under-represented in the datasets used for training~\cite{ivanovic_trajdata_2023}, which makes it hard to achieve representative human-like collision avoidance behavior solely based on learning from data~\cite{schumann_benchmarking_2023}. 

Thus, there is currently a lack of models that can capture the key aspects of human collision avoidance behavior (response selection, timing, and execution) all at the same time and across different scenarios. This limits both practical applications (due to the need to develop a new model for every new scenario) and fundamental understanding of cognitive mechanisms underlying the behavior of humans in safety-critical situations in traffic (due to the lack of a unified explanation for multiple aspects of behavior).

To address this gap, here we present a model of human collision avoidance behavior based on active inference. Originating in computational neuroscience, active inference is a versatile general framework for understanding and modeling sentient behavior in living systems~\cite{friston_active_2017, da_costa_active_2020,parr_active_2022} that has been previously used to model human behavior in diverse contexts~\cite{smith_recent_2021,vasil_world_2020,bottemanne_active_2021,harris_active_2022,novicky_precision_2024}, including the modeling of human driver behavior~\cite{engstrom_great_2018} such as car following~\cite{wei_world_2023}, responses to driving automation failures~\cite{wei_modeling_2022}, and managing uncertainty around occlusions and non-driving-related tasks~\cite{engstrom_resolving_2024}. Building on the model of Engström et al.~\cite{engstrom_resolving_2024}, our active inference model is designed to reproduce a wide spectrum of human behavior (e.g., chosen maneuver, reaction timing, collision likelihood, etc.) in response to potential collisions.
To this end, our model incorporates several well-known cognitive mechanisms to represent the dynamics of human decision making in response to sudden stimuli, such as looming perception~\cite{lee_theory_1976,gibson_ecological_2014} and evidence accumulation~\cite{luce_response_1986,gold2007neural,ratcliff2011diffusion,markkula2014modeling}. We evaluated our model against a range of previously reported empirical findings in three paradigmatic collision avoidance scenarios: 1) the front-to-rear scenario, where a driver needs to respond to a suddenly braking vehicle in front, 2) the opposite-direction lateral incursion scenario, where an oncoming vehicle suddenly cuts across the driver's path, and 3) the intersection right-turn-into-path scenario, where a vehicle coming from the right on a perpendicular road does not yield when turning right into the driver's lane. Testing the model in three markedly different scenarios allowed us to investigate the potential of active inference as a framework for generalizable modeling of human collision avoidance behavior. 

\section*{Results}
\begin{figure*}
    \centering
    \includegraphics[]{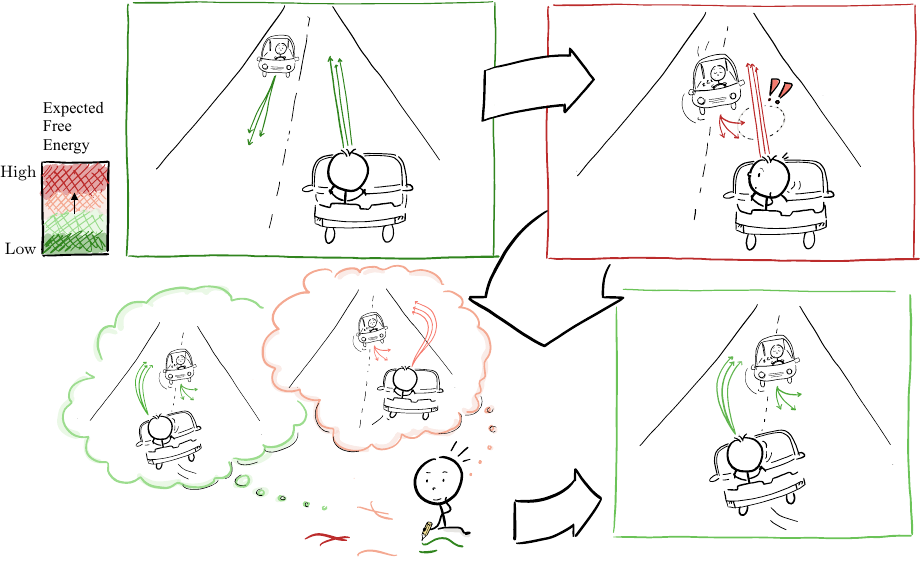}
    \caption{
    An illustration of the key principles of our active inference model in the opposite-direction lateral incursion scenario. Upper left panel: The modeled agent (at the bottom, facing forward) is continuously generating beliefs (represented by the arrows) about how the current driving situation will play out and what future observations it will make as a consequence of its own actions and the actions of the other vehicle. In "normal", non-conflict situations, the situation typically plays out as expected. The agent thus initially believes with high certainty that both itself and the oncoming vehicle will remain in their lanes and pass each other safely. This yields low expected free energy (EFE; green arrows) since the agent's expected (and thus preferred) observations (e.g., making progress, avoid collisions) are aligned with the actual observations. 
    Upper right panel: When the oncoming vehicle suddenly encroaches into the driver's lane, the EFE increases rapidly (as indicated by the red arrows) because this new situation is likely to result in an observed collision in the future, which is strongly disfavored by the agent. In this situation, the agent’s belief about the oncoming vehicle’s future trajectory also becomes more uncertain, because the other vehicle can no longer be trusted to follow established traffic norms. The agent thus needs to imagine a better future with lower EFE -- in this case, a scenario where it avoids collision by swerving left (lower left panel) and take action to make that better future come true (lower right panel).}
    \label{fig:Overview}
\end{figure*}
In this work, we use active inference as an overarching framework to guide the modeling. The key premise of active inference is that all behavior and cognition can be understood based on the single principle of minimizing free energy. An agent minimizing free energy can be conceptually understood as the agent sensing, and acting upon, the world in such a way as to minimize its surprise over time. In the active inference framework, this amounts to seeking out observations that are expected (unsurprising) and preferred given the type of creature the agent is, reflecting its adaptation to its particular environment or niche (e.g., a fish expects and prefers the sensation of being immersed in water). Agents capable of planning into the future and modeling the consequences of their actions (such as human car drivers), select plans that minimize the expected surprise, or more generally the \emph{expected free energy} (EFE)~\cite{friston_active_2017}.

Our model implements these principles in the collision avoidance context (Figure~\ref{fig:Overview}). Specifically, the modeled agent repeatedly evaluates possible futures under its current policy (i.e., a sequence of planned actions) -- including interactions with other road users -- in terms of their EFE. The driver then selects policies that minimize EFE by realizing observations that align with the driver’s preferences such as avoiding collisions (yielding \emph{pragmatic value}) while obtaining new information to reduce uncertainty in the agent's beliefs about the future (yielding \emph{epistemic value})~\cite{friston_active_2017,engstrom_resolving_2024}. 

\begin{figure*}
    \centering
    \includegraphics[width=\textwidth]{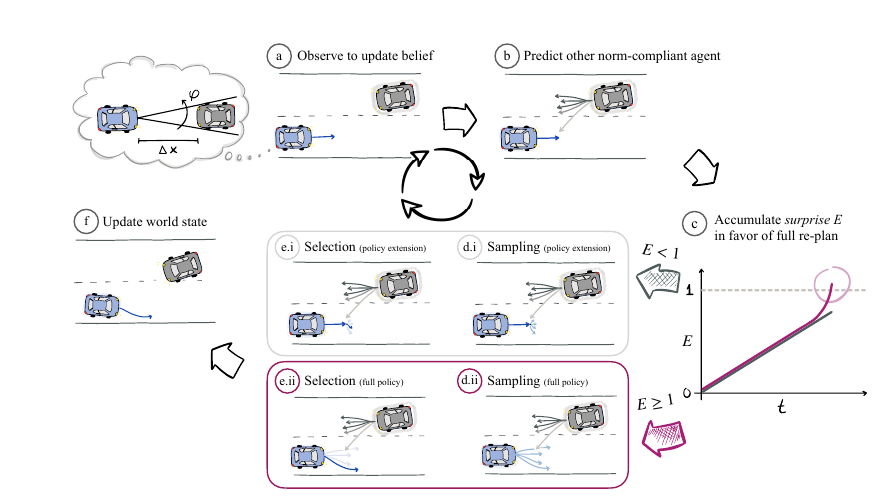}
    \caption{Main functional components of our active inference collision avoidance model. At each time step, the ego agent observes the world state, updates its belief about the world (a), and makes probabilistic predictions about the actions of the other agent (where future trajectories that adhere to traffic norms are prioritized) (b). At every time step, the agent accumulates evidence (as measured by surprise associated with the current state of the world) on the unsuitability of the current policy (c). If the accumulated evidence reaches a threshold, the agent re-plans the entire policy, selecting the one that minimizes the expected free energy (EFE) (d.ii and e.ii). Otherwise, it continues using its current policy in an extend form -- i.e., only sampling and then selecting the policy's last time step (d.i and e.i). The first action in the new or extended policy is then applied to the environment (f).} 
    \label{fig:architecture_sketch}
\end{figure*}

Fundamentally, agents cannot have perfect knowledge of the mechanisms underlying the surrounding world (the \emph{generative process}), but instead rely on an internal, not necessarily veridical, representation of their beliefs about the environment, the \emph{generative model}~\cite{friston_active_2017}. Importantly, an (ego) agent's \emph{generative model} is probabilistic, with the uncertainty about the state of the world being represented by a set of sample particles. Our model implements these general principles in a time-discrete, sequential process (Figure~\ref{fig:architecture_sketch}) which incorporates a number of key perceptual, cognitive, and motor mechanisms.

\textbf{Looming-based perception.} First, the agent observes the world and updates its belief (Figure~\ref{fig:architecture_sketch}a), combining these observations with the expectations derived from the generative model propagating its previous belief forward in time.
The model assumes that the agent cannot directly perceive kinematic states of surrounding objects (e.g., positions and velocities), but instead uses the readily available perceptual information~\cite{gibson_ecological_2014}: the object's visual angle $\varphi$ subtended at the driver's retina, and its derivative, the angular rate $\dot{\varphi}$, commonly referred to as \emph{looming}~\cite{lee_theory_1976,pekkanen_computational_2018, tian_deceleration_2022}. The model then infers back the kinematic state from those observations, where -- with increasing distances -- equal variations in $\varphi$ lead to increasingly larger variations in the inferred distance and therefore increasing uncertainty about the kinematic state of the other agent. Additionally, the agent's perception accuracy is limited: it cannot perceive looming with absolute values below a threshold $\dot{\varphi}_0$~\cite{hoffmann_estimation_1994, lamble_detection_1999}. Due to this threshold, the agent is unable to detect small relative velocities at long distances; our model thus incorporates this looming threshold as one possible mechanism behind delays in recognizing events like abrupt braking of a lead vehicle.

\textbf{Behavior prediction via norm-conditioned particle filter.} After perceiving the environment, the ego agent predicts the other vehicle's behavior (Figure~\ref{fig:architecture_sketch}b). For this, our model uses a sample-based approach~\cite{murphy2012machine, engstrom_resolving_2024} based on a particle filter. Specifically, the samples representing the agent's updated belief about the other vehicle are propagated forward by the \emph{generative model} (in our case using a bicycle model with additive Gaussian noise). In this way, the model generates multiple distinct kinematically plausible future trajectories of the other vehicle, including the ones that could potentially lead to a collision. This enables the model to anticipate rare, long-tail behaviors of the other vehicle and respond to them appropriately during collision avoidance, but yields overly pessimistic predictions in non-conflict situations. For instance, widely dispersed predictions could lead the agent to anticipate any vehicles in the adjacent lane potentially encroaching into its lane, incentivizing unnecessary evasive actions. 

Thus, relying solely on kinematic likelihood for the generation of these samples would lead to overly cautious behavior. In reality, human drivers typically must assume that other road users follow traffic rules~\cite{laurent_traffic_2021, abbas_drivers_2024} and other social norms for driving~\cite{tennant2021code} to allow for efficient interaction. To reflect this, our model's sampling process is biased towards trajectories that are not only kinematically feasible but also \emph{normatively likely}, that is, aligned with societal normative expectations on how to behave on the road~\cite{bicchieri2006rules}. We refer to this biased sampling process as a \emph{norm-conditioned particle filter}. In effect, the model’s belief about the other vehicle’s future trajectory is initially constrained by normative expectations, such as that vehicles typically remain within their lane unless there is evidence to suggest otherwise. At the same time, the normative probability is bounded from above: trajectories where norm compliance stays constant or improves over time compared to the starting state are assigned an equal normative likelihood, while those with decreasing norm compliance are deemed less likely. Thus, when another vehicle unexpectedly causes a conflict (for example, by encroaching into the ego vehicle’s lane), the model relaxes the assumption of that vehicle's future norm-compliance as well, sampling norm-violating trajectories as long as they remain kinematically plausible.

While there is generally a large set of normative expectations dependent on cultural influence, we chose to only implement specific norm-influencing behaviors in each scenario (e.g., staying in one's lane or yielding to the right of way, see Supplementary Figure~\ref{fig:Norm_prob}). A generalized, scenario-independent implementation of such norms is feasible, but with the goal of computational efficiency was not pursued here.

\textbf{Re-planning full policy based on accumulating surprise.}
In the next step, based on the set of previously predicted possible trajectories of the other vehicle, the agent considers its own policy, evaluating its current suitability by calculating the \emph{residual information}~\cite{dinparastdjadid_measuring_2023} of the \emph{pragmatic value}, which can be seen as a measure of surprise. This surprise signal -- the (scaled) negative pragmatic value associated with the currently selected policy -- essentially indicates how unsuitable the current policy is given the model's preferred observations and how the situation is expected to develop. The model assumes that the driver continuously accumulates this surprise signal as evidence in favor of full policy re-plan (Figure~\ref{fig:architecture_sketch}c). On each time step, if the accumulated evidence has not reached a predefined threshold, the agent considers it sufficient to continue with the current policy, and extends this policy one time step into the future (Figure~\ref{fig:architecture_sketch}d.i~and~e.i). However, if enough evidence in favor of current policy being unsuitable is accumulated, a set of completely new policies is sampled and a new full policy is selected (Figure~\ref{fig:architecture_sketch}d.ii~and~e.ii). The first action from either the extended or the new full policy is then applied to update the environment for the next time step (Figure~\ref{fig:architecture_sketch}f). This mechanism, where policy re-planning is contingent on the accumulation of evidence, thus provides a way to represent the dynamics of human decision making and response timing in the model.

\textbf{Constrained policy sampling.} When proposing new candidate policies, either for extension (Figure~\ref{fig:architecture_sketch}d.i) or full re-plan (Figure~\ref{fig:architecture_sketch}d.ii), the model uses the \emph{cross-entropy method}~\cite{de_boer_tutorial_2005}. With this method, candidate actions are iteratively resampled, increasingly focusing on the most promising areas of the action space. To represent humans' bounded capacity for planning under time pressure~\cite{simon1955behavioral, summala_towards_2007,oh_satisficing_2016,callaway_rational_2022, siebinga_model_2024-1}, we limit the number of evaluated policies in this process.
Importantly, the sampled acceleration values are constrained to reflect that humans operate the gas and brake pedals with one foot. These constraints include limiting applied jerks (as humans cannot press or release pedals instantaneously) and enforcing a constant holding time of $\SI{0.2}{s}$ at acceleration $a_0 \lesssim \SI{0}{ms^{-2}}$ during transitions in both directions between acceleration and deceleration (as humans cannot move their foot between pedals instantaneously).

\textbf{Policy selection via expected free energy minimization.} Among the sampled candidate policies, the model aims to find the policy that minimizes \emph{expected free energy} (EFE) -- the cornerstone of the active inference framework (Figure~\ref{fig:architecture_sketch}e.i~and~e.ii). In our model, the pragmatic value part of the EFE is maximized by 1) maintaining a desired longitudinal velocity, 2) staying within the current lane (avoiding unnecessary or unsafe lane changes) and on the road, 3) minimizing control inputs (avoid harsh braking or steering), 4) preventing collisions or at least reducing their severity (measured by the relative impact velocity), and 5) maintaining sufficient safety margins, avoiding situations where collisions may become unavoidable (such as when following a vehicle too closely without a sufficient stopping distance). Finally, maximizing epistemic value encourages 6) policies that are expected to result in a wide variety of reliable observations which may help reducing uncertainty~\cite{engstrom_resolving_2024}. However, in the present collision avoidance setting, behavior is mainly expected to be driven by pragmatic, rather than epistemic, value since there is not much the driver can do to reduce uncertainty, especially given that we here assume that the other vehicle is non-reactive to the ego agent's actions.

\subsection*{Model evaluation}
We evaluated the model against human data in three scenarios. First, in the front-to-rear scenario, we compared the model to the results of a meta-analysis of brake response times~\cite{engstroem_scenario_2010} in experimental driving simulator and field studies (e.g., Brookhuis et al.~\cite{brookhuis_effects_1991} and Lee et al.~\cite{lee_speech-based_2001}), and an analysis of deceleration magnitudes reported in~\cite{markkula_farewell_2016} based on two real-world driving datasets captured across the US (the SHRP2 dataset~\cite{victor2015analysis}) and Africa (the ANNEXT dataset~\cite{engstrom2013analysis}). 

Second, in the opposite-direction lateral incursion scenario, the model was compared to human data from a driving simulator study conducted in UK~\cite{johnson2025looking} which was conducted in parallel to the present work. Thus, by contrast to the model evaluation for the front-to-rear scenario, which was based on aggregated data, we were here able to run our model on nearly identical scenarios for which the human data was collected, allowing for more detailed comparisons. We manually fitted the model parameters to reproduce human behavior in the first two scenarios; same parameter values were used for both scenarios. 

Third, to test the ability of the model to generalize to new scenarios, we tested it, with the same parameter settings, in an unseen third scenario conducted in Canada~\cite{ziraldo_driver_2020}, an intersection right-turn-into-path scenario where another vehicle failed to yield the right of way. Similarly to the lateral incursion scenario, we were able to closely reproduce the scenarios faced by the human participants.

\subsubsection*{Front-to-rear scenario}

\begin{figure*}
    \centering
    \includegraphics[]{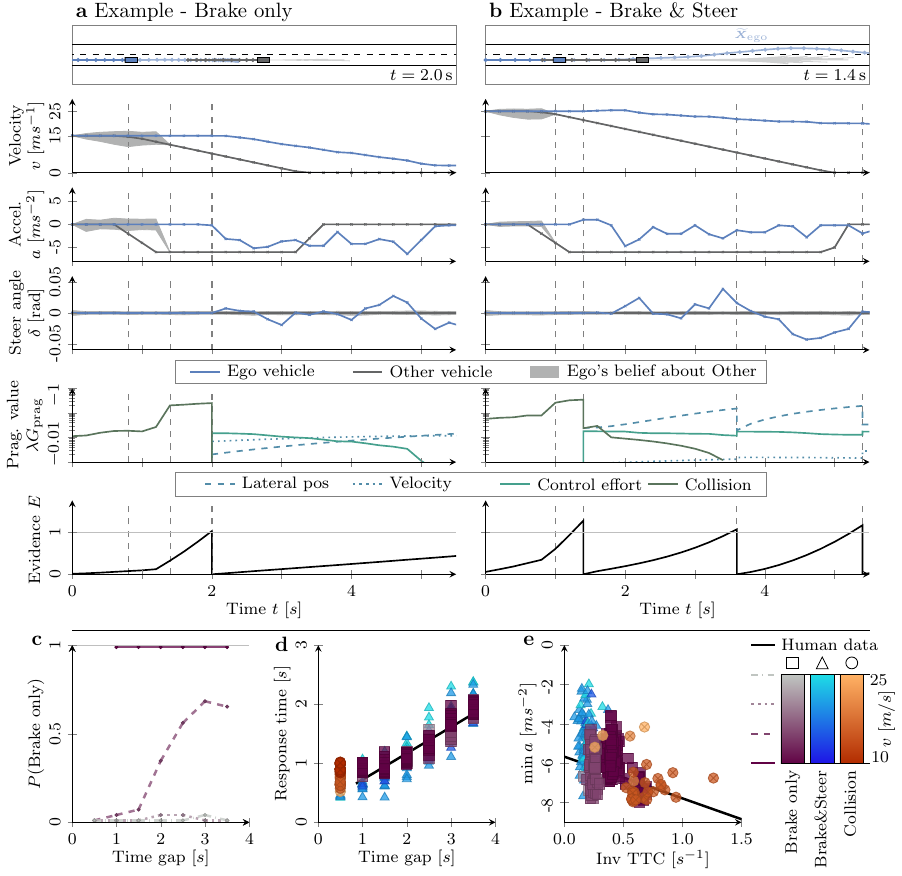}
    \caption{Evaluation of the model in the front-to-rear scenario. \textbf{a}) In a scenario where initial pre-conflict kinematics are such that collision can be avoided by braking (initial vehicle velocities $v_0=\SI{15}{ms^{-1}}$, a bumper-to-bumper time gap of $\SI{1.5}{s}$), the model typically produces braking-only behavior. The sub-panels visualize the top-down view of the scenario, the time dynamics of both vehicles' velocity $v$, acceleration $a$, steering angle $\delta$ and (for the ego vehicle) the components of the pragmatic value $G_{\text{prag}}$ (scaled with the evidence accumulation gain $\lambda$) that underlie the accumulated evidence towards a full policy re-plan. 
    \textbf{b}) In a scenario where braking on its own might not be sufficient to avoid collision ($v_0=\SI{25}{ms^{-1}}$, $\SI{1}{s}$ time gap), the model selects a policy that involves both braking and steering towards the opposite lane (see the model's selected policies in the topmost plots). \textbf{c}) The likelihood of the ego vehicle deciding to brake and stay behind the leading other vehicle (without any steering) as a function of time gap, with the four lines representing four different velocities (ranging from $\SI{10}{ms^{-1}}$ to $\SI{25}{ms^{-1}}$). This data include those samples where the agent is only steering. \textbf{d}) The relationship between the time gap and the brake response time of the ego agent in 896 model simulations (see Supplementary Information~\ref{sec:Rear_end_initial_state}). Regression line based on a meta-analysis of human brake response times~\cite{engstroem_scenario_2010} is shown for reference. \textbf{e}) The relationship of the inverse time-to-collision (TTC) at brake onset (defined as $\frac{\max\{0, v_{\text{ego}} - v_{\text{OV}} \}}{\Delta x}$) and the lowest observed deceleration. Regression line based on the human deceleration magnitude data~\cite{markkula_farewell_2016} is shown for reference. Video replays of the example simulations are available in the Supplementary Information as \emph{Supplementary\_Movie\_1.mp4} (\textbf{a}) and \emph{Supplementary\_Movie\_2.mp4} (\textbf{b}).}
    \label{fig:Rearend}
\end{figure*}

In the front-to-rear scenario, the ego vehicle trails the other vehicle which is driving in the same lane; both vehicles have the same initial velocity which was systematically varied in our simulations together with the initial time gap between the vehicles. After a short time of driving at constant speed, the other vehicle starts braking with a high constant deceleration until coming to a stop. Depending on the initial kinematics of the vehicles, the ego agent can avoid collision either solely by braking (Figure~\ref{fig:Rearend}\textbf{a}) or combined braking and swerving (Figure~\ref{fig:Rearend}\textbf{b}). The model sometimes avoided a collision by swerving only (i.e., without braking). However, since existing results in the literature on front-to-rear scenarios typically only report on braking response performance, we only compared our model's brake response times to the human response data (i.e., for the analysis of those, steering only responses were not considered).

At lower speeds and larger time gaps, the model typically avoids collisions by braking only. In a representative simulation (Figure~\ref{fig:Rearend}\textbf{a}), after the leading agent started braking at $t = \SI{0.8}{s}$ (first dashed line), it took the model $\SI{0.6}{s}$ to perceive this (see the ego agent's belief in the acceleration plot). At that point ($t=\SI{1.4}{s}$, second dashed line), the model recognized the imminent conflict resulting from this change in perceived behavior of the other vehicle, as can be seen by the sudden decrease (increase in negative value) in the collision part of the pragmatic value. This leads to a corresponding increase in the rate of accumulation of the evidence for a re-plan (i.e., surprise). However, it still took another $\SI{0.6}{s}$ until the accumulated evidence $E$ reached the threshold, delaying the first noticeable agent reaction to $t=\SI{2}{s}$ (third dashed line), where the agent started to execute a new policy -- emergency braking. This policy was selected because it allowed the agent to drastically reduce its EFE via reaching future states with high pragmatic value, in particular thanks to low collision probability. This policy was preferred over the alternative policy of braking and steering into the adjacent lane because the pragmatic value associated with lane changing was more negative than that associated with lower velocity, as the desired (corresponding to the initial) velocity is comparatively low. However, even under this policy, at $t=\SI{2}{s}$, the acceleration has still not changed, due to the pedal constraints. Consequently, the first deceleration of the model can be observed at $t=\SI{2.2}{s}$, resulting in a total response time of $\SI{1.4}{s}$.

Smaller time gaps and higher initial velocities can make it kinematically impossible to avoid collision by braking only. In such situations, the model typically opts to steer and brake at the same time (Figure~\ref{fig:Rearend}\textbf{b}). Here, due to the smaller initial distance, the model's perception delay was noticeably shorter (being only about $\SI{0.2}{s}$, with the modeled agent perceiving the braking at $t=\SI{1}{s}$, the first dashed line). However, due to the more challenging initial kinematics, it was more difficult for the model to avoid collision, despite the shorter perception delay. In particular, the model switched between different policies three times. 
Initially, the model recognized the need to slightly brake and then steer around the other vehicle: already at the time of the first re-plan ($t=\SI{1.4}{s}$, second dashed line) the collision component of EFE was substantially reduced by the chosen policy. This policy was mostly sufficient in improving the collision component of the pragmatic value, but as visible in the top panel (planned trajectory $\widetilde{\bm{X}}_{\text{ego}}$), the agent intends to stay in between lanes for an extended period. This led to two further re-plans ($t = \SI{3.6}{s}$ and $t = \SI{5.4}{s}$, third and fourth dashed lines respectively) to adjust the trajectory. 
However, even having passed the other vehicle already, the model is not completely successful, with a noticeable lateral position component remaining in the pragmatic value. As the optimal solution of driving in the center of the current lane on a free road should have maximized the pragmatic value, this remaining lateral error is likely the result of the cross-entropy method not identifying the optimal policy. This represents a bounded planning capacity which, as discussed above, is also present in humans~\cite{simon1955behavioral, summala_towards_2007,oh_satisficing_2016,callaway_rational_2022, siebinga_model_2024-1} and thus an intended feature of our model.  

By systematically varying the initial kinematics of the vehicles, we assessed the model behavior across a spectrum of front-to-rear scenarios. Due to the stochasticity of the model (foremost in behavior prediction and policy sampling), for each of the 28 sets of initial conditions we ran the simulation of the front-to-rear scenario 32 times. Based on these simulations, we analyzed the model's chosen evasive maneuver, response time, and deceleration magnitude and compared it to human data reported in the driver behavior literature.

Figure~\ref{fig:Rearend}\textbf{c} shows the probability that the model will brake only (without steering) for different velocities and time gaps. Existing studies on how human evasive maneuver choice depends on scenario kinematics in front-to-rear scenarios are limited. Thus, we compared our model results to the few related studies that exist, but these do not provide sufficiently detailed analyses for rigorously evaluating our model's evasive maneuvering decisions. 
These studies generally suggest that evasive maneuver decisions, similar to response timing, depend on the scenario kinematics (see~\cite{johnson2025looking} for a review). Avoiding collisions is generally kinematically more feasible by braking at lower speeds and by steering at higher speeds~\cite{brannstrom_decision-making_2014} and human drivers tend to swerve into an adjacent lane if a kinematically viable escape path exists~\cite{venkatraman_steer_2016, hu_decision_2017, sarkar_steering_2021, johnson2025looking}. 
In line with these general observations, our model prefers braking at lower velocities and favors combined swerving-and-braking or swerving only responses at higher velocities, regardless of the time gap (Figure~\ref{fig:Rearend}\textbf{c}). At medium velocities ($\SI{15}{ms^{-1}}$) the model behavior is sensitive to the gap between the vehicles: the longer the time gap, the higher the probability of a brake-only maneuver. This pattern reflects a trade-off between the components of the pragmatic values associated with collisions, control effort (harshness of braking and steering), and lateral position. 
In more urgent scenarios (short time gaps), the model prefers swerving due to the low pragmatic value of harsh braking. Conversely, in less urgent scenarios (larger time gaps), the model opts for braking since moderate braking avoids collision without incurring the significant reduction of pragmatic value due to leaving the lane (while the deviation from the desired velocity is still tolerable at those medium speeds). 
These specific dependencies of evasive maneuver choice on scenario kinematics have, to our knowledge, not been empirically tested and represent interesting quantitative predictions from our model. 

Regarding brake response times, empirical studies have consistently found a strong dependence on scenario kinematics~\cite{engstroem_scenario_2010, aust_effects_2013, markkula_farewell_2016, engstrom_simulating_2018, mcdonald_toward_2019, bianchi_piccinini_how_2020, engstrom_modeling_2024}. Specifically, more kinematically urgent scenarios (e.g., with a small initial time gap and/or hard lead vehicle braking) lead to shorter response times while less urgent scenarios lead to longer response times, with approximately linear relationship between measures of urgency (e.g., the initial time gap) and mean response time~\cite{engstroem_scenario_2010, engstrom_modeling_2024}.
Our model produced behavior that is remarkably consistent with these findings (Figure~\ref{fig:Rearend}\textbf{d}). Interestingly, compared to the evasive maneuver decisions, velocities do not seem to significantly influence brake response times, nor does the final chosen evasive behavior. Furthermore, model response times remain consistent (approx. $\SI{1}{s}$) over the range of short time gaps ($\SI{0.5}{s}$ to $\SI{1.5}{s}$). This last result could be a consequence of the evidence accumulation mechanism requiring certain minimum time to trigger a re-plan even at a very high rate of incoming evidence. Relatedly, the lack of fast enough responses at short time gaps is a likely explanation for the collisions observed in the model (only at the shortest time gap of $\SI{0.5}{s}$). However, because the data reported in the literature does not cover such short time gaps, this model prediction remains to be tested in future studies. 

While response selection and its timing are comprehensively characterized by the choice of evasive maneuver and response times, response execution is a dynamic process and can be characterized by a variety of metrics. Here we focus on deceleration magnitude, which has been previously shown to depend on the kinematics of the front-to-rear scenario: human drivers brake harder in more kinematically urgent scenarios~\cite{markkula_farewell_2016}.
Our model captured this phenomenon: the magnitude of its braking increased with the inverse time-to-collision at brake onset (Figure~\ref{fig:Rearend}\textbf{e}). This behavior of the model can be explained by the pragmatic value trade-off  between avoiding a collision on the one hand and avoiding the effort of hard braking and losing velocity on the other hand. In other words, the model aimed to not brake harder than necessary to avoid collision.

\subsubsection*{Opposite-direction lateral incursion scenario}\label{sec:Oncoming}
\begin{figure*}
    \centering
    \includegraphics[]{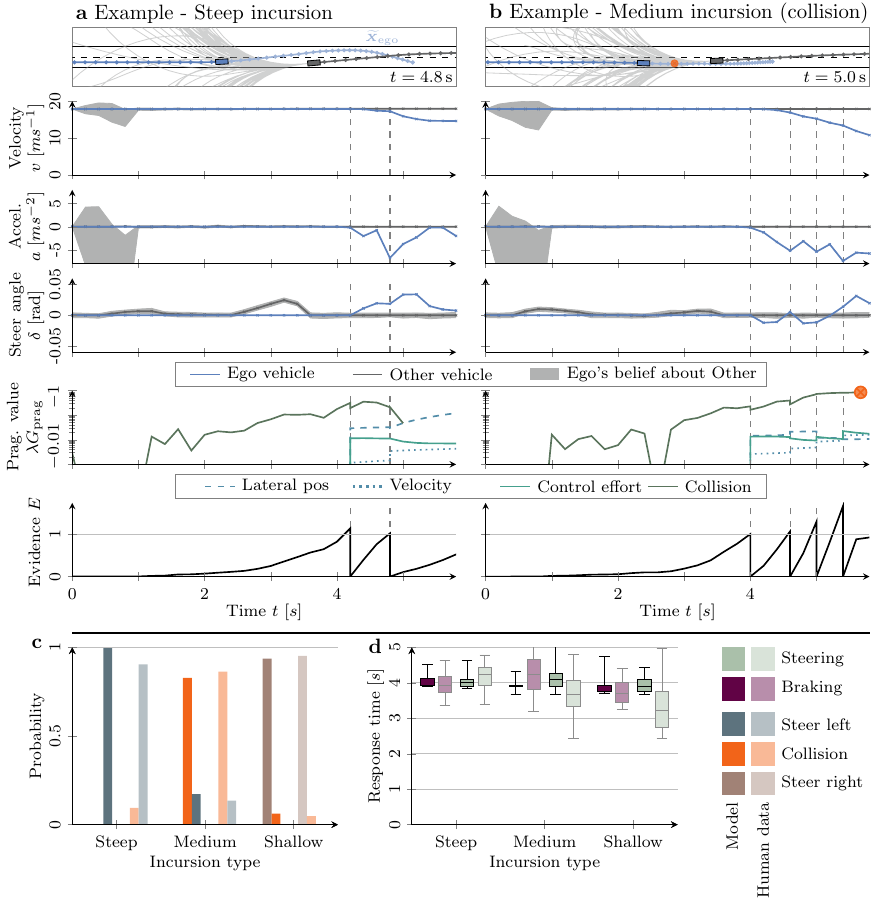}
    \caption{Evaluation of the model in the opposite-direction lateral incursion scenario. \textbf{a}) A steep, unprompted incursion by the other vehicle leads to the model swerving towards the opposite lane. The sub-panels visualize the dynamics of both vehicles' velocity $v$, acceleration $a$, steering angle $\delta$ and (for the ego vehicle) the components of the pragmatic value $G_{\text{prag}}$ (scaled with the evidence accumulation gain $\lambda$) that underlie the accumulated evidence towards a full policy re-plan. \textbf{b}) In response to a medium incursion by the other vehicle, the model attempts to avoid collision by braking and steering left, but collides at $t = \SI{5.8}{s}$ (crossed circle marker indicates the location of the collision). \textbf{c}) The likelihood of the ego vehicle successfully avoiding the other vehicle by steering left (toward the opposite lane) or right (toward the shoulder), or failing to avoid a collision. This is based on 64 model simulations and 21 to 22 human demonstrations for each of the three incursion types. \textbf{d}) Brake and steer response times. In panels C and D, human behavior and response times measured in a driving simulator study~\cite{johnson2025looking} are shown for reference. The whiskers here show the minimum and maximum observed values. Video replays of the example simulations are available in the Supplementary Information as \emph{Supplementary\_Movie\_3.mp4} (\textbf{a}) and \emph{Supplementary\_Movie\_4.mp4} (\textbf{b}).}
    \label{fig:Oncoming}
\end{figure*}

For evaluating model behavior in the opposite-direction lateral incursion scenario, we used human data from a driving simulator study at the University of Leeds, UK, reported by Johnson et al.~\cite{johnson2025looking}. In this study, vehicles initially approached each other in opposite lanes when the computer-controlled vehicle unexpectedly steered toward the participant's lane along a predefined path. The study implemented three kinematic variants differing in incursion "steepness": (1) a \emph{steep} incursion crossing in front of the participant at a sharp angle and allowing a relatively easy escape by steering toward the opposite lane; (2) a \emph{medium} incursion at a shallower angle, heading directly toward the participant; (3) a \emph{shallow} incursion only partially entering the participant's lane, making it possible to find an escape path by steering slightly toward the shoulder (the original study used left-hand traffic, so we are here avoiding directional terms to prevent confusion).

We re-implemented reversed (i.e., right-hand traffic) versions of these three conditions in our setup (see Supplementary Figure~\ref{fig:PL}), enabling a direct comparison between human and model behavior under nearly identical conditions. Each of those three conditions was repeated 64 times. 

An example of our model implemented in the steep incursion condition is shown in Figure~\ref{fig:Oncoming}\textbf{a}, while Figure~\ref{fig:Oncoming}\textbf{b} shows an example of the medium incursion condition.  
When looking at the example simulations, we can observe again that the model needs multiple re-plans before deciding on the final avoidance maneuver. Compared to the front-to-rear scenario, where this is likely caused by the model's bounded planning capacity, in the lateral incursion scenario this is due to the high uncertainty inherent in the predicted positions of the encroaching vehicle (top panels). 
In Figure~\ref{fig:Oncoming}\textbf{a} (steep incursion), a first re-planning is triggered at $\SI{4.2}{s}$ (first dashed line) but this does not significantly improve the collision component of the pragmatic value. Additionally, the model only initiates minor steering maneuvers and small accelerations/decelerations. Together, this suggests that the high uncertainty about the other vehicle's behavior prevents the agent from finding an evasive policy that is better (has lower EFE) than the current policy, instead delaying the decision, thus gaining time to see how the situation develops. With the collision risk unresolved after the first re-plan, surprise quickly accumulates toward a second re-plan at $\SI{4.8}{s}$ (second dashed line). By this time, the agents are closer, allowing less space for uncertainty to grow and enabling the model to identify a steering policy that mitigates the collision risk (after the second re-plan, the collision component of the pragmatic value no longer significantly contributes to surprise accumulation). The decision to swerve toward the left is a result of both the other vehicle having crossed sufficiently over to the right side to leave a gap for swerving, and the model’s stronger preference for moving into the opposite lane over leaving the road to the right (similar to Figure~\ref{fig:Overview}). However, it can be seen in Figure~\ref{fig:Oncoming}\textbf{a} that this trajectory is not optimal. Namely, as can be seen in the top plot, the planned trajectory $\widetilde{\bm{X}}_{\text{ego}}$ chosen during the second re-plan at $t=\SI{4.6}{s}$ (second dashed line) over-steers when returning to its own lane and consequently risks leaving the road. Thus, slight course corrections will likely be performed later (but this would occur outside the current simulation window).

Figure~\ref{fig:Oncoming}\textbf{b} shows an example of the medium incursion scenario where the model is unable to avoid a collision. Again, likely due to uncertainty about the other vehicle’s behavior, the first re-plan at $t=\SI{4.0}{s}$ (first dashed line) mainly results in a sharp deceleration and minor steering towards the right (i.e., towards the shoulder). While steering towards the right might seem suboptimal given the other agent’s current path, it aligns with the influence of the norm-conditioned particle filter, which biases the model towards believing that the other agent will likely return to its original lane. At the second re-plan ($t = \SI{4.6}{s}$, second dashed line), the model still focuses mostly on continued braking. However, this only marginally improves the collision component of the pragmatic value, leaving the imminent collision risk unresolved. At the third re-plan at $\SI{5}{s}$ (third dashed line), the agent keeps braking to reduce impact severity. At this point, the vehicles, which are still on a direct collision course with nearly no lateral offset, got so close that a collision becomes nearly unavoidable. Consequently, even the fourth re-plan ($t=\SI{5.4}{s}$, fourth dashed line) fails to find a satisfactory trajectory and thus a collision occurs shortly thereafter at $t = \SI{5.8}{s}$.

We compared our model to the human data collected in the driving simulator study by Johnson et al.~\cite{johnson2025looking}.
We closely replicated the scenarios of that study in our simulations (see Supplementary Information~\ref{sec:Oncmoing_dynamics}), which allowed direct comparisons between the model and the data with respect to the chosen avoidance behavior and both braking and swerving response times (which were extracted using the same method as in~\cite{johnson2025looking}; see the Methods section). 
A key finding of Johnson et al.~\cite{johnson2025looking} was that the participants' evasive maneuvering patterns and collision outcomes were strongly determined by the different scenario kinematics in the the three incursion scenarios. In the steep scenario, most participants avoided collisions by steering toward the opposite lane, while in the shallow scenario, participants typically steered toward the shoulder, passing the other vehicle on the "inside". However, in the medium scenario, the majority of participants collided with the oncoming vehicle. Since the urgency (i.e., time to collision at the initial steering of the oncoming vehicle) was constant across scenarios, the high collision rate in the medium case was attributed to greater uncertainty about the other vehicle’s future path, leaving no clear escape route.

As shown in Figure~\ref{fig:Oncoming}\textbf{c}, our model reproduced these results, reliably avoiding collisions in the steep and shallow scenarios through steering respectively left (toward the center) or right (toward the shoulder). Interestingly, the model also reproduced the human propensity for collision in the medium scenario. Johnson et al.~\cite{johnson2025looking} suggested that, conceptually, a key reason why human drivers tended to collide in the medium incursion scenario is the high perceived uncertainty about the oncoming vehicle's future behavior which prevents the driver from finding a sufficiently certain escape path. As described above, the current model offers a detailed computational account for the possible mechanisms underlying this phenomenon: the wide spread (uncertainty) in the behavior predictions about the oncoming vehicle, as well as a bias towards expecting that it will return to its own lane due to the norm-conditioning of the beliefs, prevents the model to find an evasive policy in time to avoid collision  (see Figure~\ref{fig:Oncoming}\textbf{b} above). 
An additional factor behind the observed collision rates was likely also the model's bounded planning capacity. Increasing the model's planning capacity -- by increasing the number of evaluated policies in the cross entropy method tenfold -- resulted in a significant drop in the model's collision rate (from 82.3\% to 51.6\% in the medium incursion scenario, and from 6.3\% to 3.1\% for shallow incursions, for further discussion see Supplementary Information~\ref{sec:CEM_parameters}).

Furthermore, Johnson et al.~\cite{johnson2025looking} found that braking and steering response times -- measured from initiation of the incursion to the first reaction -- were generally long, ranging around $\SI{3.5}{s}$ to $\SI{4.5}{s}$. 
While braking response times did not vary noticeably with incursion level, steering response did decrease consistently from the steep towards the shallow scenario.
Our model generally reproduced the range of response times, with closely matched median values (Figure~\ref{fig:Oncoming}\textbf{d}), but did not capture the variation in steering response times (especially for the shallow incursion, the model typically steered at least half a second later). 
One possible explanation may be that many participants avoided collision in the steep scenario primarily by braking before making a final steering adjustment, delaying their recorded steering response. In contrast, the model relied more on steering for collision avoidance, requiring an earlier response. 
The underlying reasons for these behavioral differences between the model and human drivers are still unclear but may have to do with varying individual preferences for braking versus swerving. Further work is needed to see if these individual differences can be reproduced by the model by varying its velocity, lane change, and lateral control effort preferences.

\subsubsection*{Intersection scenario}

\begin{figure*}
    \centering
    \includegraphics[]{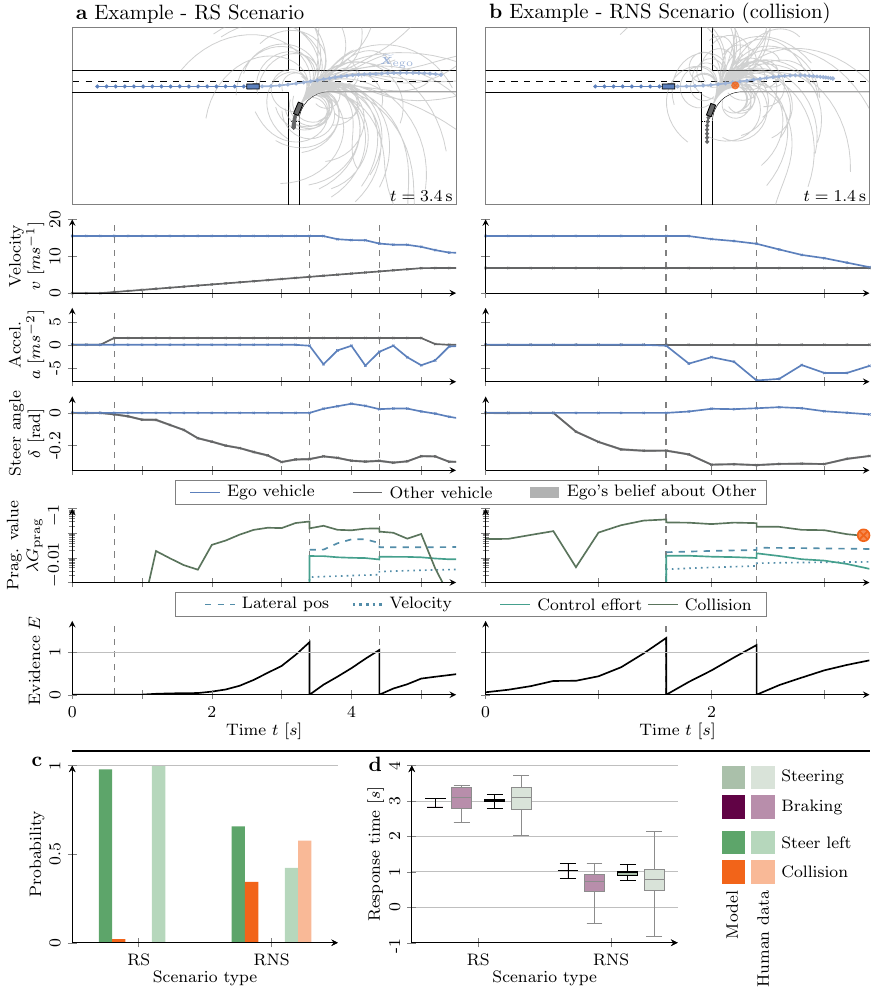}
    \caption{Evaluation of the model in the intersection scenario. \textbf{a}) Right-turn-stopped (RS): A vehicle initially waiting at the stop line performs a right turn on red without yielding; by accelerating and turning into the ego vehicle's lane it forces the model to steer left into the parallel lane to avoid a collision. The sub-panels visualize the dynamics of both vehicles' velocity $v$, acceleration $a$, steering angle $\delta$ and (for the ego vehicle) the components of the pragmatic value $G_{\text{prag}}$ (scaled with the evidence accumulation gain $\lambda$) which constitutes the accumulated evidence towards a full policy re-plan. \textbf{b}). Right-turn-not-stopped (RNS): Another agent approaches the intersection with a constant speed, making a right turn into the model's lane without yielding or even slowing down, leading to a collision at $t = \SI{3.4}{s}$ (crossed circle marker indicates the location of the collision). \textbf{c}) The likelihood of the ego vehicle successfully avoiding the other vehicle (by steering to the left or braking) or colliding with it, based on 96 model simulations and 26 human demonstrations for each of the two scenario variants. \textbf{d}) Brake and steer response times. In panels C and D, human event outcomes and response times obtained in a driving simulator study~\cite{ziraldo_driver_2020} are shown for reference. The whiskers here show the minimum and maximum observed values. Video replays of the example simulations are available in the Supplementary Information as \emph{Supplementary\_Movie\_5.mp4} (\textbf{a}) and \emph{Supplementary\_Movie\_6.mp4} (\textbf{b}).}
    \label{fig:Intersection}
\end{figure*}

To test the ability of our model to generalize to more complex scenarios and to unseen data, we evaluated it in an intersection scenario against human data that was not used for tuning model parameters. Specifically, we used human response data obtained in a driving-simulator study conducted in Guelph, Canada, by Ziraldo et al.~\cite{ziraldo_driver_2020} in the context of collision avoidance at intersections. This study implemented a right-turn-into-path scenario, where the participants approached an intersection with the right of way (green light). A computer-controlled vehicle then approached from the right side, on an intersecting road, and performed a right turn on red into the participant's lane. Right turns on red are allowed in US and Canada but the turning vehicle must yield; the participant's vehicle therefore had the right of way. Two variations of this scenario were implemented: In the right-turn-stopped (RS) scenario variant, the other vehicle initially waited at the stop line in front of the intersection and began to accelerate when the modeled agent’s time-to-arrival was approximately $\SI{4}{s}$. In the right-turn-not-stopped (RNS) scenario variant, the other vehicle was moving towards the intersection at a constant speed, and was controlled such that, under a constant-velocity assumption, it would cross the stop line when the modeled agent was roughly $\SI{1.5}{s}$ from entering the intersection. In both scenarios, there was a parallel lane to the left of the modeled agent, allowing it to steer left to avoid a collision. A detailed version of our implementation of these scenario variants can be found in Supplementary Information~\ref{sec:intersection_setup}. While Ziraldo et al.~\cite{ziraldo_driver_2020} had each of their 52 participant face three of those scenarios, we only consider the first responses (26 per scenario type), as significant changes in response times and collision rates between repetitions indicated an anticipation of the conflict scenario. In total, we ran each of those scenario variants 96 times, varying the initial velocity of the modeled agent within the limits specified by Ziraldo et al.~\cite{ziraldo_driver_2020}.

In the RS scenario (Figure~\ref{fig:Intersection}\textbf{a}), after the other vehicle begins to accelerate at $t=\SI{0.6}{s}$ (first dashed line), several time steps pass before the model considers the collision risk to be significant (as reflected in the collision penalty component of the pragmatic value). As evidence (i.e., surprise) thereafter gradually accumulates, the model reaches the re-planning threshold at $t=\SI{3.4}{s}$, selecting a policy that combines braking with a leftward steering adjustment. However, as there is still considerable uncertainty about the other vehicle’s future actions -- particularly since the other vehicle is already violating normative expectations by not yielding, reducing the bias introduced by norm conditioning -- the inferred collision risk decreases only slightly. This necessitates an additional re-planning step one second later (third dashed line). Shortly afterward, the ego vehicle passes the other vehicle, resolving the conflict.

In the RNS scenario (Figure~\ref{fig:Intersection}\textbf{b}), the situation becomes critical earlier, as the other vehicle is approaching at constant speed. Since the vehicle does not exhibit any observable braking to yield at the stop line, as would be normatively expected, the model considers it increasingly likely that the other vehicle will encroach into its path. As a result, surprise accumulates based on the increasing collision likelihood. At $t=\SI{1.6}{s}$ (first dashed line), the ego vehicle initiates braking and applies a slight leftward steering adjustment. 
However, this response is largely insufficient to reduce the collision risk (the collision component of the pragmatic value does not change substantially). While the model generally predicts the other agent to make a right turn, the high uncertainty in this prediction -- given the norm violations by the other vehicle, it is mostly based on kinematic likelihood -- leaves no clear safe escape path. The reaction is also too late for braking (although it is attempted) to prevent a collision. Additional re-planning at $t=\SI{2.4}{s}$ (second dashed line) likewise fails to avert the collision ($t=\SI{3.4}{s}$). Had the model opted for a more extreme steering maneuver at the first re-planning point, it could have avoided the collision. However, in this specific simulation, the added pragmatic value of such an extreme maneuver was unclear at that point due to the high uncertainty about the other agent's future trajectory. 
Nevertheless, as the policy sampling process is stochastic, the model can occasionally opt for early steep steering and thus avoid collisions in this scenario.

Compared the model to human responses in the same scenario, we found that our model closely reproduced the main observations reported in~\cite{ziraldo_driver_2020}. As in the human data, the model successfully avoided most collisions in the RS scenario, but often collided in the RNS scenario (Figure~\ref{fig:Intersection}\textbf{c}). The slightly lower collision rate of the model in the RNS scenario might be explained by the fact that in this scenario approximately ten percent of human participants collided without any noticeable reaction, which might indicate distraction effects that are not represented in our model. Next to collision rate distributions, our model reproduced the variation in human reaction times in the two scenarios (Figure~\ref{fig:Intersection}\textbf{d}): responses in the more critical RNS scenario were markedly faster than in the RS scenario in both human data and model simulations. The model closely matched the timing of human responses in both conditions, with median reaction times generally not being apart more than $\SI{0.2}{s}$. This is especially noteworthy, as it was achieved in an unseen scenario without any parameter tuning. This indicates that the model can generalize to new scenarios and can account for human data that was not used for parameter fitting/tuning. While empirically, this generalizability is limited to interactions with a single target agent, given the strong support for the underlying model components in the literature, it can be hypothesized that the model's generalizability extends further, although this has to be evaluated in future work.
Similar to the lateral incursion scenario (Figure~\ref{fig:Oncoming}\textbf{d}), the variances of model's reaction times are smaller compared to the human data. However, this is most likely simply a consequence of using a single model with fixed parameters for all simulations, thereby ignoring potential individual differences between human participants.

\subsection*{Evaluating individual model mechanisms}

\begin{figure*}
    \centering
    \includegraphics[]{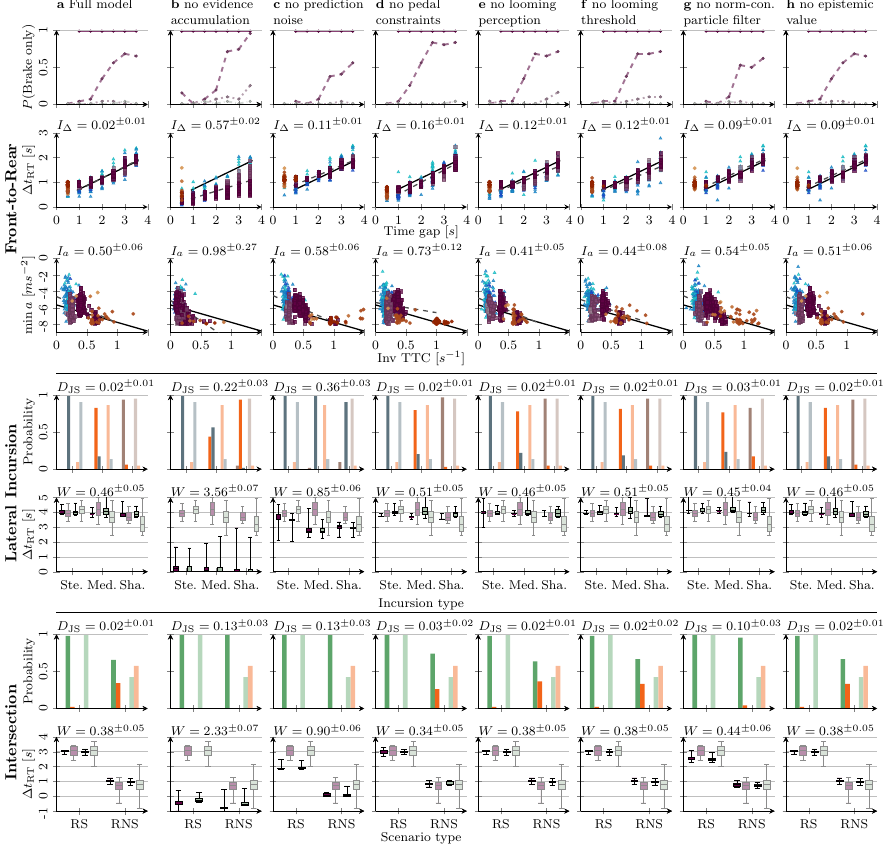}
    \caption{Comparing the full model (\textbf{a}) to variants of the model that systematically exclude its key mechanisms (the columns \textbf{b}-\textbf{h} refer to different model ablations). We consider the front-to-rear scenario (top three rows, see Figure~\ref{fig:Rearend} for the legend), the lateral incursion scenario (middle two rows, see Figure~\ref{fig:Oncoming} for the legend), and the intersection scenario (bottom two rows, see Figure~\ref{fig:Intersection} for the legend). In the second and third rows, dashed gray lines depict the \emph{maximum a posteriori} linear regression of the model simulation results, with the corresponding mean absolute error $I$ between that line and the ground truth ($I_\Delta$ is given in $s$, while $I_a$ is given in $ms^{-2}$). In the fourth and sixth row, we used the Jensen-Shannon divergence $D_{\text{JS}}$ and in the fifth and seventh row the Wasserstein distance $W$ (in $s$) to quantify the difference between the model and empirical distributions. For all metrics, expected values $\pm$ standard deviation are indicated.}
    \label{fig:Ablation}
\end{figure*}
Comparison to the empirical data (Figures~\ref{fig:Rearend},~\ref{fig:Oncoming},~and~\ref{fig:Intersection}) revealed that our model captured the key aspects of human collision avoidance performance in three different collision avoidance scenarios. To better understand which of the model mechanisms are essential for capturing the empirical observations, we evaluated the performance of seven simpler models which systematically excluded the key mechanisms one-by-one (Figure~\ref{fig:Ablation}).

\textbf{Removing evidence accumulation}. Without evidence accumulation, the model fully re-plans its policy at every time step. Consequently, it starts reacting to changes in the other vehicle's behavior immediately after detecting them. This resulted in a drastic reduction in response times across all scenarios (Figure~\ref{fig:Ablation}\textbf{b}). 
Additionally, the model without evidence accumulation predominantly failed to avoid collisions in the shallow incursion scenario, contrasting sharply with human data and the full model. This was mostly caused by the model initiating braking very early. Given the uncertainty about the other vehicle, neither steering left nor steering right allows safe escape paths, so the agent continues braking, coming to a complete stop. At the point in time where the certainty about the other agent is low enough for safely choosing either left or right, the agent can physically no longer avoid a collision (as it cannot move laterally when stopped). An example simulation of this scenario can be found in the Supplementary Information as \emph{Supplementary\_Movie\_7.mp4}. The fact that collisions happened mostly in the shallow incursion scenarios is likely because in the medium and steep incursion scenarios, premature braking leaves enough space in front of the ego agent for the other vehicle to move across, which is not the case in the shallow incursion. In this scenario, a collision would still occur if the ego agent stayed in its starting lateral position, which makes braking an insufficient avoidance mechanism.
Finally, in the intersection scenario, immediate reaction (including both braking and steering) allowed the reduced model to avoid all RNS collisions, in contrast to both the full model and the human data. 
Based on those observations, the inclusion of evidence accumulation is critical for accounting for empirical data on human response times in all three scenarios, evasive maneuver selection in the shallow lateral incursion scenario, and collision rates in the intersection scenario.

\textbf{Removing added noise in behavior prediction}. The model that uses deterministic predictions about the other vehicle produced response times in the front-to-rear scenario that are only slightly longer than for both the full model and the human data (Figure~\ref{fig:Ablation}\textbf{c}), especially given shorter time-gaps. By contrast, in the opposite-direction lateral incursion and intersection scenarios, removing the noise led to substantially shorter response times. The reason for this apparent discrepancy is that in the front-to-rear scenario, prediction noise results in ego agent's believing that the other vehicle may brake more aggressively, prompting the model to react earlier. In the other scenarios, however, removing the prediction noise decreases the ego agent's uncertainty about the other vehicle, which enables it to find a safe escape path earlier, leading a faster response. 

In the lateral incursion scenario specifically, this leads to earlier steering responses, which, given how these particular scenarios unfolded, turned out be a more effective avoidance strategy, as evidenced by the absence of collisions in this case. However, this strategy was only more effective after the fact (i.e., with hindsight of how the scenario played out). In another counterfactual scenario, where the oncoming vehicle turns back into its own lane early, swerving could instead result in a collision (see~\cite{johnson2025looking} for a discussion of this important point). 
Prediction noise affected not only response times but also the model's choice of evasive maneuver in the lateral incursion scenario. In particular, the model with deterministic predictions predominantly opted to avoid the other vehicle by steering left regardless of the incursion profile. This allowed the model to avoid collisions in all simulations, but is not consistent with the human data. 

Similarly, in the intersection scenario, this model ablation successfully avoided all collisions as well. Taken together, these findings underscore that behavior prediction noise is crucial for accurately reproducing human behavior, particularly in the lateral incursion and intersection scenarios. Specifically, uncertainty in the predictions of the other vehicle's actions seems to be a key explanation for why most human drivers ended up in a collision in the medium lateral incursion scenario and the RNS intersection scenario (as discussed above in relation to Figure~\ref{fig:Oncoming}\textbf{b} and Figure~\ref{fig:Intersection}\textbf{b}, respectively).

\textbf{Removing pedal constraints}. The model not constrained by pedal switching delays resulted in approximately $\SI{0.2}{s}$ faster brake response times in all three scenarios compared to the full model (Figure~\ref{fig:Ablation}\textbf{d}). This matches the delay of $\SI{0.2}{s}$ implemented to represent the switch of the driver's foot from the gas pedal to the brake pedal.
Although the effect appears simple and straightforward, the implemented pedal constraints are essential for capturing the differences between steering and braking response times observed in human behavior. For example, in the lateral incursion scenario, our model still slightly overestimates steering response times while underestimating braking response times, a discrepancy that could not be corrected by adjusting the surprise threshold for re-planning, as this would shift all response times uniformly.

\textbf{Removing looming perception.} We analyzed a version of the model which directly perceived kinematic variables without relying on looming (Figure~\ref{fig:Ablation}\textbf{e}) as well as one which perceived looming but did not have a looming perception threshold (Figure~\ref{fig:Ablation}\textbf{f}).  
In the front-to-rear scenario, both these models differed from the full model and the human data in that they produced faster brake response times at longer time gaps. For the model without the looming threshold, this is due to the fact that greater time gaps (and therefore greater distances) require a larger velocity difference to surpass the looming detection threshold, which delays responses, as it needs more time to accumulate. Furthermore, these results being very similar to the model that did not perceive looming at all (Figure~\ref{fig:Ablation}\textbf{e}) indicates that in this scenario, the looming threshold was a key factor behind the ability of the model to capture human response times. Additionally, while not specifically analyzed here, this effect is expected to be exacerbated in high speed freeway scenarios where (given the higher speed) the following distances for a given time gap increase, further emphasizing the importance that the model accounts for this phenomenon. In the other two scenarios, removing the looming threshold (or looming perception entirely) did not have any major effects, as the relevant observations in those scenarios are based on lateral movements.

In summary, models without either looming perception or looming threshold resulted in qualitative differences between the model and human data on brake response times in the front-to-rear scenario. In addition to the strong theoretical support, this highlights the importance of both looming-based perception and looming perception threshold for scenarios with largely longitudinal dynamics.

\textbf{Removing the norm-conditioned particle filter}. Without the norm-conditioned particle filter, the model behavior in the front-to-rear scenario changes only slightly compared to the full model (Figure~\ref{fig:Ablation}\textbf{g}), showing marginally longer response times. In the lateral incursion scenario, however, the effect is more pronounced, with collisions being less likely in the medium incursion (avoided by leftward steering) and more likely in the shallow incursion. Relative to the full model, this ablation lacks the bias against predicting steeper incursions earlier, which incentivizes steering towards the left. This yields a small advantage in the medium incursion scenario, where avoiding towards the other vehicles lane is typically effective, but increases collision risk in the shallow incursion, where both the full model and humans consider rightward avoidance the better option. 
Additionally, the ablated model also shows slightly longer reaction times in the lateral incursion scenario. This again follows from the lack of norm conditioning, as without it, the model often predicts that the other vehicle will leave the road before posing a collision risk, whereas the full model -- biased by the normative filter -- believes the other vehicle will stay in its lane. 
Consequently, in this ablation, with the predicted collision likelihood being lower, surprise accumulates more slowly, and the re-planning threshold is reached later, leading to a slower reaction.
In the intersection scenario, removing the norm-conditioned particle filter leads to both faster reaction times and a successful resolution of the conflict in nearly every instance. This is logical, as the expectation that the other vehicle would stop to yield resulted in delayed reaction. Without this bias, potential collisions are taken more seriously and reacted to earlier.

While these differences already highlight the importance of the norm-conditioned particle filter, it has the strongest impact in benign (non-conflict) scenarios. In such scenarios, the model with the simple kinematics-based behavior prediction mechanism is unrealistically conservative due to having to plan for all kinematically possible futures of the other vehicle --- a problem that is avoided by the model with norm-conditioned particle filter (see Supplementary Information~\ref{sec:Benign} for an example). 

\textbf{Removing the epistemic component of EFE.} 
Lastly, we investigated a model variant that ignored the epistemic value when selecting policies. This ablation, with the exception of the reaction times in the front-to-rear scenario, did not result in any significant differences in the model behavior, both in the actual outcome and the derived goodness-of-fit metrics. Consequently, it can be argued that the EFE part of our model could potentially be reduced to pragmatic value alone, which would make the model conceptually similar to a belief-space model predictive control~\cite{van2012motion}.
Nevertheless, the epistemic value does play a role in scenarios where, for instance, observability is limited; there, epistemic value has substantial effect on model behavior (Supplementary Information~\ref{sec:Epistemic}). This complements the previous investigations of the role of epistemic value in scenarios requiring visual behavior~\cite{engstrom_resolving_2024}. For these reasons, we treat the version of the model with epistemic value as the ``full'' model here to allow for generalization to similar scenarios in the future.

Additionally, in the lateral incursion scenario, we examined the impact of our model's bounded reasoning capacity. Specifically, as noted above, increasing the number of policies evaluated during replanning led to a substantial reduction in observed collision rates—ranging from roughly one-third to one-half across scenarios. This suggests that bounded reasoning is a crucial component for accurately modeling human collision-avoidance failures. However, since not all collisions were avoided, other factor (namely, prediction uncertainty and the expectation of norm-compliant behavior) also play a significant role in driving such outcomes.

Overall, the seven ablated model variants analyzed above show qualitative differences compared to the full model, and none reproduce human behavior across all three scenarios as well as the full model. This provides further evidence for the importance of all the analyzed model mechanisms in regard to capturing human collision avoidance behavior.

\section*{Discussion}
Building on previous work~\cite{engstrom_resolving_2024}, we here presented an active inference-based framework for modeling human collision avoidance behavior based on first principles.
To our knowledge, this is the first published computational model that offers a detailed account of human collision avoidance behavior across multiple conflict scenarios. Our model reproduces many key results in the rich literature on collision avoidance in front-to-rear scenarios, in particular the dependence of response times and braking magnitude on scenario kinematics. The model also generated several detailed predictions on how evasive maneuver decisions (braking vs. swerving) vary with scenario kinematics which can be tested in future experiments. Furthermore, we compared the model to detailed human behavior data in an opposite-direction lateral incursion scenario with varying kinematics~\cite{johnson2025looking} and an intersection scenario with with a right-turning vehicle failing to yield the right of way~\cite{ziraldo_driver_2020}, both obtained in driving simulator studies. This analysis demonstrated that the model can reproduce human evasive maneuver decisions, collision outcomes, as well as response times in different variations of these two scenarios. The model also shows strong generalization across scenarios, with the model tuned solely on the front-to-rear and lateral incursion scenario making accurate predictions of human responses in the intersection scenario.

%The driving simulator study in~\cite{johnson2025looking} provides a rich source of detailed human collision avoidance data that the model can be evaluated against but, due to space constraints, we only report the main findings here. Further work could further explore the extent to which the model also reproduces more detailed aspects of the data, for example response sequences and acceleration patterns. 

%%%%%%%%%%%%%%%%%%%%%%%%%%%%%%%%%%%%%%%%%%%%%%%%%%%%%%%%%%
%                   Collision avoidance modelling        %
%%%%%%%%%%%%%%%%%%%%%%%%%%%%%%%%%%%%%%%%%%%%%%%%%%%%%%%%%% 

A key feature of our proposed model is its ability to flexibly represent closed-loop collision avoidance through dynamic re-planning of policies, including braking, steering, and accelerating actions, in a way that is qualitatively and quantitatively similar to human collision avoidance behavior. In contrast, existing publicly available collision avoidance models either implement a "one-shot", often predetermined, open-loop evasive maneuver~\cite{bargman_how_2015, kusano_safety_2012, olleja2025validation} or limited aspects of closed-loop control, such as intermittent braking but no steering~\cite{svard_quantitative_2017, svard_computational_2021}. Some commercially available models -- the Stochastic Cognitive Model (SCM) from BMW~\cite{fries_driver_2022, fries_modeling_2023} and the driveBOT model from cogniBIT~\cite{de_oliveira_simulation-based_2023,rossert_cognitive_2024} -- reportedly implement human-like closed-loop behavior in collision avoidance scenarios. However, although the SCM is openly available, the published evaluations provide only a limited basis for meaningful comparison. For instance, Fries et al.~\cite{fries_modeling_2023} evaluates SCM on only a few collision-avoidance cases and compares mainly velocity-over-time profiles, without a systematic analysis of diverse metrics across scenario variations. For driveBOT, the evaluations are more comprehensive~\cite{de_oliveira_simulation-based_2023,rossert_cognitive_2024}, but the lack of available information on the detailed working principles of the model precludes a meaningful comparison with our work.

Our model achieves closed-loop behavior by continuously evaluating evasive policies based on their expected free energy, pursuing the policy with the lowest EFE, and allowing for further re-planning if the chosen policy turns out to not yield the preferred (and hence expected) outcomes (e.g., due to limited planning accuracy or unexpected changes in how the scenario unfolds). This, in effect, implements a constraint satisfaction mechanism where different model preferences are traded against each other in finding the policy with the highest overall combined pragmatic value and epistemic value. We observed several examples of this (Figures~\ref{fig:Rearend}) where, for example, the model generally prefers to brake at low speeds (since the loss related to the preferred speed is not as high as the cost of changing lane) and swerve at higher speeds (where the cost of braking in terms of speed loss are higher). This yields detailed predictions of human behaviors in different kinematic situations that could be further tested empirically.
 
Another important aspect of our model is its ability to dynamically account for uncertainty about the future behavior of other agents. This is particularly important in situations like the lateral incursion scenario and the intersection scenario. For instance, in the former, the driver cannot determine \emph{a priori} if the other vehicle will continue crossing the ego vehicle's lane or if it will turn back to its original lane, resulting in a wide range of potential future trajectories of the other vehicle. However, the uncertainty may be reduced as the scenario unfolds, opening up new available escape paths, or \emph{escape affordances}~\cite{johnson2025looking}, as was the case in the current steep and shallow incursion scenarios. In such cases, an evasive response that is initiated too early may be premature (and non-optimal) and does not properly reflect human behavior. We saw an example of this in Figure~\ref{fig:Ablation}\textbf{c}, where in both the lateral incursion and intersection scenario the model without prediction noise responded much faster than humans and the full model, thus avoiding collision in all cases. However, such overconfident evasive maneuvering behavior could be detrimental in other scenarios, for instance if in our lateral incursion scenario the other vehicle would turn back into its original lane. By representing the uncertainty about the other vehicle's future positions using a vehicle dynamics (bicycle) model and a noisy particle filter, our model is able to handle such situations and successfully reproduce human behavior.

Furthermore, in particular the intersection scenario showed that it is also important that such noisy predictions are biased towards norm-compliant behavior (Figure~\ref{fig:Ablation}\textbf{g}). Specifically, without the prior expectation that other vehicle will follow established rules and norms such as yielding the right of way, the model will predict potential collision scenarios earlier, react faster and resolve the conflict safely. However, in many situations this leads to overreactive non-human like collision avoidance behavior. We also demonstrated that the \emph{norm-conditioned particle filter} is necessary to avoid overly conservative (and therefore non-human like) behavior in non-conflict driving scenarios (Supplementary Information~\ref{sec:Benign}).

In the collision avoidance scenarios simulated in the present study, there is little opportunity for the driver to reduce the uncertainty about the oncoming vehicle's future trajectory through epistemic actions. In other words, there is little the driver can do to obtain further information that could help predict what the other vehicle will do. Thus, as expected, disabling the epistemic value component of the EFE did not significantly change the model behavior. However, as shown in Supplementary Information~\ref{sec:Epistemic} and previous work~\cite{engstrom_resolving_2024}, the epistemic value component in our model does drive policy selection in traffic scenarios with epistemic affordances, that is, situations with opportunities for actions yielding information that may resolve uncertainty. Hence, even if the epistemic value did not have any significant effect on policy selection in the present study, it should still be considered a key aspect of human road user behavior. %as further discussed below, 
In particular, in the real world communicative acts such as honking, flashing headlights or initially moving to the right in the lane could reduce uncertainty about the other vehicle driver's intent. Since other agents were here assumed to be non-reactive, such aspects were not addressed in the current work; this is a topic for future development of the model.

While active inference models can be implemented computationally in many different ways, we used a stochastic receding-horizon control architecture as the basis for the implementation, following~\cite{engstrom_resolving_2024}. This shares similarities with prior work in robotics, particularly decision-theoretic approaches that explicitly account for uncertainty in actuation and sensing by propagating belief states into the future, often using particle-based representations, and then selecting actions that optimize expected outcomes based on this projected belief evolution~\cite{hubmann2017decision, brechtel2014probabilistic, bouton2017belief}. However, our work diverges from these engineering models in two crucial aspects. First, since our goal is to represent human driver behavior rather than designing a robot/vehicle controller, the modeling focuses on capturing key aspects of human driving, such as response timing and evasive maneuver decision making, rather than optimizing performance based on engineering criteria. Second, our modeled agent minimizes EFE (as prescribed by the active inference framework), while traditional state-based reward functions often employed in the control literature lack such a theoretically-grounded objective derived from first principles. 

Our model incorporates an explicit mechanism for evidence accumulation to account for human response timing, based on existing models~\cite{tian_deceleration_2022,pekkanen_variable-drift_2022, zgonnikov_should_2022, schumann_using_2023, engstrom_modeling_2024}. Evidence accumulation models naturally account for the situation-dependency of response timing in traffic situations: the faster a traffic conflict escalates, the faster the human's response to it~\cite{markkula_farewell_2016, engstrom_modeling_2024, zgonnikov_should_2022}. Compared to existing evidence accumulation models, our model introduces some key novel aspects. In particular, in contrast to traditional models that accumulate perceptual evidence (e.g., looming~\cite{tian_deceleration_2022}), our model rather accumulates surprise. Similar ideas have been explored in existing models~\cite{bianchi_piccinini_how_2020, svard_quantitative_2017,dinparastdjadid_measuring_2023,engstrom_modeling_2024}, based on the notion that a traffic conflict by definition is an extremely rare (and hence surprising) event which always takes place against a "default" expectation of how the situation would normally play out (Figure~\ref{fig:Overview}). This is also the key idea behind the NIEON (Non-Impaired driver with their Eyes ON the conflict) response time model which conceptualizes the onset of the stimulus that human drivers respond to in a traffic conflict as the onset of surprise (i.e., the violation of the initial default expectation)~\cite{engstrom_modeling_2024}. 
The current model generalizes this idea in several ways. First, instead of triggering a predefined action such as braking when the surprising evidence has reached the threshold, the evidence accumulation here triggers the~\emph{re-planning} of a new policy, using the EFE-based policy selection mechanism discussed above. Second, the surprise signal that is being accumulated is not just the difference between a predicted and actual signal but rather the negative pragmatic value of the current policy, that is, how "bad" the currently selected default policy is relative to the agent's preferred observations. This provides a straightforward solution to the known problem of how to determine whether a given surprise signal is relevant to the agent~\cite{dinparastdjadid_measuring_2023}. 

Importantly, our model accumulates surprise deterministically without accumulation noise. Under this assumption, the accumulation rate can be expressed in units of the decision boundary (another key parameter traditionally important in the evidence accumulation literature, interpreted as response caution) and hence we did not need to define a separate decision boundary~\cite{ratcliff2016diffusion}. 
Nevertheless, adding such accumulation noise could be seen a future improvement of our model if we want to better match the high variance observed in human reaction times (Figure~\ref{fig:Oncoming}\textbf{d} and Figure~\ref{fig:Intersection}\textbf{d}). Similarly, the addition of surprise decay could also be considered.

While we here model evidence accumulation explicitly, it can in principle be seen as an implicit feature of active inference~\cite{fitzgerald_active_2015}. However, in practice, it is challenging to accurately represent human response dynamics implicitly in a simplified model like ours. To enable straightforward tuning of the model's response performance to human data, we chose the current explicit modeling of evidence accumulation. As demonstrated by our simulation results (Figure~\ref{fig:Ablation}\textbf{b}), the inclusion of this mechanism is essential to account for realistic response times in our model.

%%%%%%%%%%%%%%%%%%%%%%%%%%%%%%%%%%%%%%%%%%%%%%%%%%%%%%%%%%
%                   Active inference discussion          %
%%%%%%%%%%%%%%%%%%%%%%%%%%%%%%%%%%%%%%%%%%%%%%%%%%%%%%%%%%
The current model is based on the general model predictive control architecture originally developed for the model of routine driving described in~\cite{engstrom_resolving_2024} which, in turn, is based on the standard discrete-time active inference EFE model for planning agents formulated as a partially observable Markov decision process~\cite{friston_active_2017}. The original routine driving model in~\cite{engstrom_resolving_2024} focused primarily on the role of epistemic value in resolving uncertainty when driving around visual occlusions and during visual time sharing (drawing from the rat-in-a-T-maze example in~\cite{friston_active_2017}). However, that model was not aimed at describing time-critical behaviors. For instance, in collision avoidance scenarios its responses to sudden stimuli would be overly fast, similar to the version of our model without evidence accumulation (Figure~\ref{fig:Ablation}\textbf{b}). Extending the original model~\cite{engstrom_resolving_2024} with not only evidence accumulation, but also other mechanisms such as looming-based perception and norm-conditioned predictions was essential for capturing human collision avoidance behavior (Figure~\ref{fig:Ablation}). Otherwise, the model we reported here retains the key features of the original routine driving model needed for dealing with uncertainty resolution through epistemic action, thus offering a powerful and general computational framework for modeling driving behavior, also beyond collision avoidance. As in~\cite{engstrom_resolving_2024}, we employed engineering tools like the cross-entropy method and particle filters with discrete and continuous variables. However, while relying on such established methods, our model does not reduce to a combination of them; instead, these methods serve as means for implementing specific cognitive mechanisms within our active inference framework.

Our framework is grounded not only in active inference and the free energy principle~\cite{parr_active_2022}, but also enactivist approaches to cognitive science~\cite{varela2017embodied, thompson2010mind}. Active inference and enactivist cognitive science share the the foundational notion that behavior in biological agents results from a self-organizing process geared towards sustaining the agent's integrity (existence) over time. This emphasizes a strong continuity between life and mind~\cite{thompson2010mind,friston_life_2013} and offers a natural solution to the problem how the certain aspects of the environment become meaningful and relevant to an agent~\cite{kiverstein_problem_2022}. Such an existential imperative can be described as the organism striving to maintain itself in a sparse attractive set of preferred agent-environment states, where the attractive set depends on the specific organism in question. Thus, to survive over time, a fish needs to remain in water with the right chemical balance and temperature range, obtain food, avoid being eaten by certain predators etc. This implies that these particular states are meaningful and relevant to the fish which then acts in order to observe these states with a high degree of certainty, thus looking as if it is minimizing free energy over time. By the same token, we endowed our model with a (highly simplified) set of preferred states that are meaningful and relevant to a human car driver (avoid danger of collision, stay on the road, keep up progress, avoid too harsh braking/steering) and defined the model such that it acts (selects and executes policies) in a way that maximizes the chance of observing these preferred states.  

Related to this, our model also offers a operationalization of the classical notion of \emph{affordance}, traditionally broadly defined by Gibson~\cite{gibson_ecological_2014} as what the environment ``offers the animal, what it provides or furnishes, either for good or ill'' (p. 127). Thus, when our model identifies candidate evasive policies associated with low EFE, these can be understood as affordances in the sense of opportunities for actions that are worth pursuing. The active inference framework offers a further distinction between~\emph{pragmatic affordances} yielding unsurprising, familiar and preferred observations and~\emph{epistemic affordances} yielding information that can resolve uncertainty about pragmatic affordances~\cite{friston_affordance_2022, engstrom_resolving_2024,ramstead_easy_2022} (see ~\cite{johnson2025looking}) for a further discussion about this expanded notion of affordances and its relation to the classical affordance concept in ecological psychology).

Thanks to generality of active inference, models based on it have been developed for virtually every facet of biologically-based cognition and behavior, in organisms ranging from single cells~\cite{friston_life_2013}, plants~\cite{calvo_predicting_2017}, individual animals and humans~\cite{van_de_maele_hierarchical_2024} to social and cultural phenomena~\cite{kastel_small_2023}, which provides a great source of concepts and ideas for modeling different aspects of human road user behavior. Thus, importantly, we see active inference not just as a computational approach to model development, but as a general theoretical framework for understanding road user behavior that can guide modeling at a conceptual level and generate hypotheses for experimental studies. At the same time, the great majority of existing active inference models use relatively simple toy scenarios, thereby limiting its applicability to real-world human behavior. By operationalizing active inference in the context of a dynamically rich task, that is, collision avoidance and evaluating it against human data from both simulated and real-world driving, this study pushes the boundaries of active inference applications toward the modeling of complex human naturalistic behaviors. 

%%%%%%%%%%%%%%%%%%%%%%%%%%%%%%%%%%%%%%%%%%%%%%%%%%%%%%%%%
%        Limitations                                    %
%%%%%%%%%%%%%%%%%%%%%%%%%%%%%%%%%%%%%%%%%%%%%%%%%%%%%%%%%

While, as discussed above, our model represents a significant advancement in human collision avoidance modeling, it has several limitations that can be addressed in future work. First, in this work we essentially hand-tuned the model parameters to fit the human data from the two collision avoidance scenarios. Whereas this highlights the model's generalizability and interpretability, especially considering its good performance on the third (intersection) scenario not used for tuning, such an approach is impractical at scale. Future research should explore systematic parameter optimization methods, for example using the approach suggested by Wei et al.~\cite{wei2025learning}.

Second, the current model assumes that other agents are non-reactive (i.e., in the generative model's state transition function, other agents' states are independent from each other and the ego agent). This assumption therefore limits the model to scenarios where interactions are only weakly linked, and is unlikely to hold in the general case. For example, norms about priority -- such as at unsignalized intersection -- are generally interactive, with one agent’s norm compliance depending on the other agents’ behavior. Thus, future work should extend the model to include reactive agents by implementing interactive norms and communicative or epistemic actions that allow agents to probe or signal intentions. Such extensions would strengthen the model's applicability and allow simulations of multi-agent interactions under active inference. 
Moreover, the current model (as well as the original routine driving model in~\cite{engstrom_resolving_2024}) only reasons at a shallow "unsophisticated" level about how actions bring about future observations with varying degrees of uncertainty. Further work could explore the possibilities of incorporating \emph{sophisticated inference}~\cite{friston2021sophisticated}, which enables "deeper" levels of recursive reasoning about how future observations would lead to updated beliefs and corresponding new actions. This may be particularly relevant when modeling interactions between road users.

Third, the model currently only accounts for perceptual limitations in terms of visual angle and looming. However, looming only represents the relative motion of objects moving longitudinally towards the observer and, strictly speaking, only applies to objects located along a straight path of travel relative to the gaze direction of the observer (in our case, this means objects located straight ahead of the ego vehicle). Thus, additional perceptual variables are needed to represent perceptual thresholds on the detection of perpendicular motion (e.g., a pedestrian moving into the street ahead) or representing relative motion of objects that are otherwise perpendicular from the observer’s gaze and heading direction. In addition, the perceptual effects of environmental and stimulus-related factors such as weather (e.g., reduced visibility due to fog, heavy rain or snow), lighting conditions (daylight, dusk, night, effects of headlights etc.), glare and stimulus conspicuity would ideally need to be accounted for. Another way in which the model’s perceptual fidelity could be improved is by explicitly modeling the human field of view, including the different functional properties of foveal versus peripheral vision which can be combined with the visual behavior model outlined in the original model~\cite{engstrom_resolving_2024}. 

Beyond perceptual limitations, the use of more realistic perceptual variables such as the optically specified longitudinal time-to-collision ($\tau$ in~\cite{lee_theory_1976}), bearing angle (specifying perpendicular collision course~\cite{chohan2006postural}) and global optic flow rate or edge rate (specifying self motion relative to the ground~\cite{larish1990sources,fajen2005calibration}) could be further explored. This would allow for defining observation and state spaces in the model which are more aligned with the ecological information available to human drivers. 

Fourth, while mechanisms for dealing with partial observability during occlusions and visual time sharing were explored in~\cite{engstrom_resolving_2024}, further work is needed to develop more comprehensive solutions for representing and reasoning about multiple agents hidden behind occluding objects or appearing outside the driver's field of view (see also Wei et al.~\cite{wei_navigation_2024}); this could help translate the model to more complex and crowded conflict scenarios.

Fifth, the model’s generalizability beyond the presented scenarios from the Anglo-American cultural context has not been evaluated. Three types of assumptions are especially noteworthy: (1) the low-level control assumption that drivers use the same foot for acceleration and braking, (2) the higher-level assumptions about other peoples behavior encoded in normative likelihoods (for example, the different rules regarding right turns on red light in North America and Europe), and (3) the own preferences defining acceptable behavior (such as infringing on other agents' lanes). As those assumptions are subject to cultural influences~\cite{pele2017cultural, ozkan2006cross}, they may not transfer seamlessly to other regions of the world. 
A future model could be made more robust by making such assumptions explicit and learnable (e.g., by modeling them as additional latent (belief) states or context variables dependent on a country).

Finally, while the current policy sampling and kinematic-based particle filter approach were demonstrated to work well in the current scenarios, this could be a limiting factor in scaling the model to a wider range of scenarios, especially when modeling pre-conflict scenarios with more complex road infrastructures and road user interactions. To overcome this limitation, machine-learning-based data-driven models~\cite{salzmann_trajectron_2020, mangalam_goals_2021, girgis_latent_2022, shi_motion_2022, nayakanti_wayformer_2023} could be leveraged to generate possible futures in our generative model's state transition function, either stand-alone or in combination with the current kinematics-based predictions, thus potentially further enhancing the generalizability of the model. 
In principle, such data-driven models, if employed in a closed-loop manner, could potentially serve as an alternative to our proposed active inference model in its entirety, given their clear advantages in inference time -- no expensive policy selection process is needed -- as well as their superior ability to learn from large human datasets in complex geometric environments. These characteristics, after all, allowed those models to show remarkable, and increasing accuracy in the prediction of routine driving behavior~\cite{schumann2025step}. However, research on evaluating such models in collision avoidance scenarios is lacking. To understand their current capabilities, we tested one of the state-of-the-art machine-learned models (ADAPT~\cite{aydemir_adapt_2023}) in the lateral-incursion scenario (Supplementary Information~\ref{sec:Oncoming_ML}). We found that the data-driven model did not capture human behavior well, likely due to simplistic learning objectives and under-representation of critical scenarios in the training data. Furthermore, the black-box nature of such models restricts them to an exclusively descriptive/predictive role, without providing insight into the underlying cognitive mechanisms driving human behavior.

Overall, our work contributes to the recent body of research~\cite{wei2025learning,engstrom_resolving_2024, wei_active_2024, wei_navigation_2024} with new evidence that active inference can serve as a general framework for computational road user modeling. Along with the emerging real-world applications of active inference in robotics~\cite{lanillos_active_2021} and artificial intelligence~\cite{mazzaglia_free_2022}, this work can aid in the development of agents used for the simulation-based evaluation of advanced driver assistance systems and automated vehicles~\cite{fremont2020formal,montali_waymo_2023,uzzaman2025testing}.

\section*{Methods} \label{sec:methods}
\subsection*{Model principles}
Our model follows the architecture of the original model initially formulated by Engström et al.~\cite{engstrom_resolving_2024}, and combines some of the original elements of that model with multiple new mechanisms. In particular, our model retains the representation of probabilistic beliefs using sets of samples and the correspondingly sample-based estimation of the expected free energy (see equation~\eqref{eq:pragmatic_actual}) as well as the cross entropy method used for policy selection. In addition to these mechanisms, we introduced looming perception, evidence accumulation, the norm-conditioned particle filter, and the constraints on acceleration profiles (see Figure~\ref{fig:Ablation} for the analysis of the impact of these new mechanisms on model behavior). We also modified the Bayesian belief update (equation~\eqref{eq:variation_inference_orig}) and the calculation of the epistemic value (equation~\eqref{eq:epistemic_actual}).

In our active inference model, the environment is represented by the \emph{generative process}, with a state of the world $\bm{\eta} \in \mathcal{E}$ which is partially observable by the agent. The agent can influence the environment with its actions $\bm{a} \in \mathcal{A}$ (according to the state transition probability $\widehat{p}(\bm{\eta}'\vert \bm{\eta}, \bm{a})$), resulting in observations $\bm{o} \in \mathcal{O}$, whose dependence on the world state is described in $\widehat{p}(\bm{o}\vert \bm{\eta})$ (the $\widehat{p}$ is used to show these function be part of the \emph{generative process}). 

The agent has an internal model of the world, the \emph{generative model}, with the corresponding partially observable state $\bm{s} \in \mathcal{S}$, and the corresponding state transition function $p(\bm{s}'\vert \bm{s}, \bm{a})$ and observation probability $p(\bm{o}\vert \bm{s})$. As the \emph{generative model} is an abstraction of the real world described by the \emph{generative process}, $\bm{\eta}$ and $\bm{s}$ do not necessarily have to represent the same state spaces ($\mathcal{E} \neq \mathcal{S}$). The agent is uncertain about the state of the generative model, represented by the probability density function $q(\bm{s})$ (referred to as the agent's belief about the state $\bm{s}$). Following~\cite{engstrom_resolving_2024}, in our model the agent represents its belief distributions $q(\bm{s})$ in a non-parametric way as a set of $N$ samples $\bm{S} = \{\bm{s}_1, \hdots, \bm{s}_N\}$ (i.e., a particle filter).

The agent then tries to minimize its free energy, both in the belief $q(\bm{s})$ it forms (\emph{variational free energy}) and the actions $\bm{a}$ it chooses (\emph{expected free energy}). The action and policy selection is based on the existence of a preference prior $p(\bm{o})$ depicting preferred observations.

\subsection*{Perception}
%%% 1. Base perception
At time $t$, the agent perceives the world state $\bm{\eta}_t$, resulting in the observation $\bm{o}_t$: 
\begin{equation}
    \bm{o}_t \sim \widehat{p}(\bm{o}'\vert \bm{\eta}_{t})
\end{equation}
With this observation, the agent updates its old belief $q(\bm{s}_{t-1})$ to $q(\bm{s}_{t})$.
The active inference framework postulates that this update is based on the minimization of the \emph{variational free energy}~\cite{friston_active_2017}, but in our case we use standard Bayesian updating:
\begin{equation}\begin{aligned} \label{eq:variation_inference_orig}
    q(\bm{s}_{t}) & \propto p(\bm{o}_{t}\vert \bm{s}_{t})\, \mathbb{E}_{q(\bm{s}_{t-1})} p(\bm{s}_{t}\vert \bm{s}_{t-1}, \bm{a}_{t-1})\,.
\end{aligned}\end{equation}
Here, the particle filter first advances the individual samples in $\bm{S}_{t-1}$ (representing $q(\bm{s}_{t-1})$) using the transition function to get the expected states $\bm{S}_{A, t}$  (approximating $q_A(\bm{s}_{t}) = \exp \left(\mathbb{E}_{q(\bm{s}_{t-1})} \ln p(\bm{s}_{t}\vert \bm{s}_{t-1}, \bm{a}_{t-1})\right)$), with $\bm{s}_{A, t, n} \sim p(\bm{s}'\vert \bm{s}_{t - 1, n}, \bm{a}_{t-1}) $ being randomly sampled. Then, a kernel density estimate~\cite{deisenroth_mathematics_2020, fischer_information_2020} based on $\bm{S}_{A, t}$ is used to approximate $q_A(\bm{s}_{t})$ as a Gaussian Mixture Model (GMM). Assuming that $p(\bm{o}_{t}\vert \bm{s}_{t})$ is Gaussian as well, $q(\bm{s}_{t})$ can also be represented as a GMM; this allows straightforward sampling of $\bm{S}_{t}$. Compared to the approach in the original model~\cite{engstrom_resolving_2024} of weighting samples $\bm{s}_{A,t,n} \in \bm{S}_{A, t}$ with $p(\bm{o}_{t}\vert \bm{s}_{A,t,n})$ and then resampling based thereon, this approach allowed us to represent large shifts in beliefs outside the extreme values of the initial set $\bm{S}_0$.

%%% 2. Looming-based perception
The agent's visual perception in our model is based on looming (Supplementary Information~\ref{sec:Looming}) --
the optical information immediately available to humans about the relative motion of objects straight ahead in the direction of travel~\cite{gibson_ecological_2014}. Specifically, the relative position and motion of objects along the forward path of the agent is perceived in terms of the visual angle $\varphi$ subtended by the object at the retina of the observer, and its derivative, the angular rate $\dot{\varphi}$ (looming)~\cite{lee_theory_1976,pekkanen_computational_2018}. Due to the nonlinear relationship between the observer's distance to the observed vehicle (in Figure~\ref{fig:architecture_sketch}\textbf{a} referred to as $\Delta x$) and $\varphi$ (with increasing distances, equal variations $\delta \Delta x$ lead to increasingly smaller $\delta \varphi$), perception errors for $\varphi$ and $\dot{\varphi}$ will lead, respectively, to increasing inference errors and therefore uncertainty about the position and speed of the leading other vehicle at increasing distances. We represent this aspect in the model by setting the noise in $p(\bm{o}\vert \bm{s})$ to be constant across $\varphi$ and its derivatives. 

%%% 3. Looming threshold
At a certain distance, $\dot{\varphi}$ is no longer perceptible to the human eye, which implies a minimal threshold for looming detection~\cite{hoffmann_estimation_1994,lamble_detection_1999}. To account for this, the model incorporates a detection threshold on $\dot{\varphi}$ (i.e., for small $\vert\dot{\varphi}\vert$, the means of the observation probabilities $p(\bm{o}\vert \bm{s})$ for velocities and acceleration were set to the values corresponding to $\dot{\varphi} = 0$). This means that changes in the other vehicle's velocity cannot be perceived by a human driver beyond a certain distance, resulting for example in a delayed recognition if the lead vehicle suddenly brakes. 

\subsection*{Behavior prediction}
After observing the current state of the world, the agent applies the state transition function $p(\bm{s}'\vert \bm{s}, \bm{a})$ starting from its current belief $q(\bm{s}_t)$ in order to generate a predicted belief about the future states of the situation $\widetilde{q}_s(\bm{s}_{\tau}\vert \bm{\pi}_t, q(\bm{s}_t))$ (with $\tau \in \{t+1, \hdots, t+H\}$, an arbitrary policy $\bm{\pi}$ containing the ego agent's sequence of planned actions, and prediction horizon $H$) of all the agents in the scene, as well as the corresponding belief over possible future observations $\widetilde{q}_o(\bm{o}_{\tau}\vert \bm{\pi}_t, q(\bm{s}_t)) = \mathbb{E}_{\widetilde{q}_s(\bm{s}_{\tau}\vert \bm{\pi}_t, q(\bm{s}_t))} p(\bm{o}_{\tau}\vert \bm{s}_{\tau})$. During this prediction process, the agent adds noise to the control inputs of the other agents to represent the possibility that they might change their behavior.

Practically, the agent advances the belief $\widetilde{q}_s(\bm{s}_{\tau}\vert \bm{\pi}_t, q(\bm{s}_t))$ (represented by $\widetilde{\bm{S}}_{\tau} = \{\bm{s}_{\tau,1}, \hdots, \bm{s}_{\tau,N}\}$) to the next time step by using the state transition function (in our case based on a bicycle model~\cite{polack_kinematic_2017}, see Supplementary Information~\ref{sec:state_transition}) on each of the sample states $\bm{s}_{\tau,n} \sim p(\bm{s}_{\tau} \vert \bm{s}_{\tau - 1,n}, \bm{a}_{\tau - 1})$ (similarly to forming the expected belief $q_A(\bm{s})$ in the belief update). The belief $\widetilde{q}_o(\bm{o}_{\tau}\vert \bm{\pi}_t, q(\bm{s}_t))$ is then represented by the set $\widetilde{\bm{O}}_{\tau} = \{\bm{o}_{\tau,1}, \hdots, \bm{o}_{\tau,N}\}$, where $\bm{o}_{\tau,n} \sim p(\bm{o}_{\tau}\vert \bm{s}_{\tau,n})$. 

This belief propagation scheme represents \emph{unsophisticated inference} in the technical sense that it does not condition the beliefs about future states on corresponding counterfactual observations. While it would theoretically be possible to use a form of \emph{sophisticated inference}~\cite{friston2021sophisticated} that accounts for such conditioning, we here used the unsophisticated approach for simplicity, similar to the original model~\cite{engstrom_resolving_2024} and most existing active inference models.

Importantly, our state transition function $p(\bm{s}'\vert \bm{s}, \bm{a})$ function is composed out of two parts, with
\begin{equation}
    p(\bm{s}'\vert \bm{s}, \bm{a}) \propto \breve{p}_{\text{n}}(\bm{s}') p_o(\bm{s}'\vert \bm{s}, \bm{a}) \,.
\end{equation}
Here, $p_o(\bm{s}'\vert \bm{s}, \bm{a})$ is the standard kinematic likelihood function (as used in~\cite{engstrom_resolving_2024}), while the \emph{projected normative probability} $\breve{p}_{\text{n}}(\bm{s}')$ biases the agent to predict future states which currently, and over the near future, are aligned with the norm-compliant behavior of other vehicles. This represents the human tendency to avoid overly pessimistic predictions about the behavior of other traffic participants by instead assuming that they will adhere to road rules and other established traffic norms~\cite{laurent_traffic_2021, abbas_drivers_2024}; for example, assuming that oncoming vehicles will stay in their lane or yield the right of way, without evidence to the contrary, is necessary for fluent driving. 

In practice, we define $\breve{p}_{\text{n}}$ with
\begin{equation}
    \breve{p}_n(\bm{s}_\tau) \propto \min \left\{ p_{\text{n}}(\bm{s}_\tau), 2\frac{p_{\text{n}}(\breve{\bm{s}}_{\tau+1})\,p_{\text{n}}(\breve{\bm{s}}_{\tau+H_{\text{n}}})}{p_{\text{n}}(\breve{\bm{s}}_{\tau+1}) + p_{\text{n}}(\breve{\bm{s}}_{\tau+H_{\text{n}}})}   \right\}\, ,
\end{equation}
applying the \emph{normative probability} $p_{\text{n}}$ to the current time as well as short- and medium-term predictions $\breve{\bm{s}}$ ($H_n$ is setting the medium-term prediction horizon) to evaluate their norm compliance. This ensures that the model's predictions are not only biased against actions which lead to immediate norm violations, but also against those that put the other vehicle into a position where norm compliance becomes kinematically less likely over time (e.g., initiating a turn while in a straight lane does not immediately lead to the vehicle leaving the road, but might put the vehicle in a position where staying on the road becomes kinematically impossible).
Crucially, the use of the current norm-compliance as an upper bound ensures that once norm-violating behavior is observed, the model no longer places trust into norms as a way of constraining predictions about other agents' behavior. In collision avoidance scenarios initiated by the unexpected, norm-violating behavior of other agents, this allows the model to account for all kinematically possible but low-probability ``extreme'' (long-tail) behaviors. 

This approach -- which we call \emph{norm-conditioned particle filter} -- helps aligning the model's predictions with realistic driving scenarios while accounting for norm violations when necessary. 
Even though the norms relevant for our studied scenarios (lane following, yielding the right of way) could in general be implemented globally based on map information (for example a detailed graph), for simplicity and computational efficiency we chose to implement in each scenario only those norms that are relevant to that scenario (Supplementary Information~\ref{sec:Normative_Belief_F2R},~\ref{sec:Normative_Belief_ODLI}~and~\ref{sec:Normative_Belief_Intersection}; see also Supplementary Information~\ref{sec:state_transition} for details on sampling from $p(\bm{s}'\vert \bm{s}, \bm{a})$).

%%%
\subsection*{Policy sampling} 
\label{sec:Policy_sampling}
% Importance sampling
Based on the predicted behavior of the other vehicle, the agent selects its policy $\bm{\pi}$. Here, policies are defined as sequences of $H$ future actions; following the bicycle model, the actions are defined in terms of acceleration $a_{\text{long}}$ and steering rate $\omega$. These actions are selected using the cross-entropy method~\cite{de_boer_tutorial_2005, engstrom_resolving_2024}, which iteratively resamples sets of $M$ candidate policies $K$ times, where the resampling focuses around the subset of the $\beta M$ best samples from the previous iteration. Specifically, the agent takes the mean and variance of this promising subset (ignoring potential correlations), and then samples a new set with size $M$ from the corresponding normal distribution. In contrast to the original model~\cite{engstrom_resolving_2024} that randomly picked the final sampling output from the last iteration's set, here we specifically choose the best sample.

% Constraints
To better represent human behavior, the model incorporates a pedal constraint to filter out unrealistic changes in acceleration (Supplementary Information~\ref{sec:Control_limits}). As human drivers generally control acceleration by operating the brake and gas pedal with a single foot and cannot move the foot between pedals instantaneously, the model imposes a constant holding time interval of \(\SI{0.2}{s}\) at \(a_0 \lesssim \SI{0}{ms^{-2}}\) (similar to~\cite{weber_towards_2023}), such that if the model transitions from a current acceleration \(a_{\text{long}} > a_0\) to a target acceleration \(a_{\text{long},f} < a_0\), it must first decelerate to \(a_0\), maintain that value for \(\SI{0.2}{s}\), and then proceed to \(a_{\text{long},f}\). The same applies for acceleration changes in the other direction. 

\subsection*{Policy roll-out and evaluation}
% Basic idea of generative planning
To evaluate a policy $\pi$, the agent first uses the bicycle model to roll out its possible future states based on $\bm{\pi}$. In the general case, the possible future states of the agent and the other vehicle are coupled, hence the agent rolls out its policies jointly with the predicted futures of the other vehicle. For the reason of simplicity, however, our model assumes that in $p(\bm{s}'\vert \bm{s}, \bm{a})$ the other vehicle is unresponsive to the actions of the ego vehicle, allowing us to separate behavior prediction and policy roll-outs.

% Policy evaluation
Active inference then rests on the fundamental assumption that an agent prefers policies which minimize its expected free energy (EFE) $G$ over the prediction horizon of $H$ time steps; in other words, the best policy is the one with the lowest EFE. This EFE combines the desire of the agent to, on the one hand, observe itself in states that it prefers (maximize pragmatic value $g_{\text{pragm}}$) and, on the other hand, seek information that may reduce uncertainty about the world (maximize epistemic value $g_{\text{epist}}$):
\begin{equation}\begin{aligned}
    G(\bm{\pi}_t, q(\bm{s}_t)) = \sum\limits_{\tau = t + 1}^{t + H}  & - g_{\text{pragm}}\left(\widetilde{q}_o(\bm{o}_{\tau}\vert \bm{\pi}_t, q(\bm{s}_t))\right) \\ & - g_{\text{epist}}\left(\widetilde{q}_s(\bm{s}_{\tau}\vert \bm{\pi}_t, q(\bm{s}_t)) \right).
\end{aligned}\end{equation}
The pragmatic values $g_{\text{pragm}}$ is defined as
\begin{equation}
    g_{\text{pragm}}\left(\widetilde{q}_o(\bm{o})\right) = \mathbb{E}_{\widetilde{q}_o(\bm{o})} \ln p(\bm{o}) \label{eq:pragmatic}  \, ,
\end{equation}
while we use the Shannon entropy $\mathcal{H}$ to calculate the epistemic value~\cite{friston_active_2017}
\begin{equation}\begin{aligned}
    g_{\text{epist}}\left(\widetilde{q}_s(\bm{s}) \right) & = \mathbb{E}_{\widetilde{q}_o(\bm{o})} D_{KL}\left[\widetilde{q}(\bm{s}\vert \bm{o})  \vert \widetilde{q}_s(\bm{s})  \right]    \\
    & = \mathcal{H}(\mathbb{E}_{\widetilde{q}_s(\bm{s})} p(\bm{o}\vert \bm{s}))  - \mathbb{E}_{\widetilde{q}_s(\bm{s})} \mathcal{H}(p(\bm{o}\vert \bm{s})) \, . \label{eq:epistemic}
\end{aligned}\end{equation}
The first term in Equation~\eqref{eq:epistemic} is known as the~\emph{posterior predictive entropy} and represents the extent to which a policy is expected to yield a variety of different observations. The second term, known as \emph{expected ambiguity} represents the diversity of observations expected for a given state, that is, the extent to which the observations of that state are unreliable (e.g., due to reduced visibility).

In our particle-based representation, we represent $\widetilde{q}_s(\bm{s})$ with the set $\widetilde{\bm{S}}$ of $N$ particles and $\widetilde{q}_o(\bm{o}) = \mathbb{E}_{\widetilde{q}_s(\bm{s})} p(\bm{o}\vert \bm{s})$ with the set $\widetilde{\bm{O}}$, resulting in the approximations 
\begin{equation}
    \tilde{g}_{\text{pragm}}\left(\widetilde{\bm{O}}\right) = \frac{1}{N} \sum\limits_{\bm{o} \in \widetilde{\bm{O}}} \ln p(\bm{o})\label{eq:pragmatic_actual}
\end{equation}
and
\begin{equation}\begin{aligned}
    \tilde{g}_{\text{epist}}\left(\widetilde{\bm{S}},\widetilde{\bm{O}}\right) = & \, - \frac{1}{N} \sum\limits_{\bm{o} \in \widetilde{\bm{O}}} \ln \left( \frac{1}{N} \sum\limits_{\bm{s} \in \widetilde{\bm{S}}}  p(\bm{o}\vert \bm{s}) \right) \\ & \, - \frac{1}{N} \sum\limits_{\bm{s} \in \widetilde{\bm{S}}} \mathcal{H}\left(p(\bm{o}\vert \bm{s}) \right) \, , \label{eq:epistemic_actual}
\end{aligned}
\end{equation}
under the assumption that the entropy $\mathcal{H}\left(p(\bm{o}\vert \bm{s}) \right): \mathcal{S} \rightarrow \mathbb{R}$ can be calculated analytically. The original model~\cite{engstrom_resolving_2024} had calculated $g_{\text{epist}}$ by using KDE methods based on $\widetilde{\bm{O}}$ to approximate $\widetilde{q}_o(\bm{o}) = \mathbb{E}_{\widetilde{q}_s(\bm{s})} p(\bm{o}\vert \bm{s})$, compared to us using $\frac{1}{N} \sum\limits_{\bm{s} \in \widetilde{\bm{S}}}  p(\bm{o}\vert \bm{s})$. We opted for this approach, as the original estimation would result in an over-approximation of $\widetilde{q}_o(\bm{o})$, given that we evaluate the KDE exclusively at the center of each kernel. Consequently, we use the following approximation $\tilde{G}$ for the EFE:
\begin{equation}\begin{aligned}
    \tilde{G}(\bm{\pi}_t, q(\bm{s}_t)) = \sum\limits_{\tau = t + 1}^{t + H} & - \tilde{g}_{\text{pragm}}\left(\widetilde{\bm{O}}_{\tau}\right) \\ 
    & - \tilde{g}_{\text{epist}}\left(\widetilde{\bm{S}}_{\tau},\widetilde{\bm{O}}_{\tau}\right).
\end{aligned}\end{equation}

When evaluating the pragmatic value, we assume that the preference (i.e. the distribution of desired observations) $p(\bm{o})$ accounts for four aspects of the agent's objective: a) to maintain its desired longitudinal velocity, b) to minimize the magnitude of control inputs, c) to remain on the road and within its current lane (e.g., avoiding lane markers or opposite lanes), d) prevent collisions, and e) avoiding hazardous situations. Each of these aspects is represented by an individual preference function (Supplementary Information~\ref{sec:Preferences}), which are multiplied together:
\begin{equation}\label{eq:preference_likelihood}
    \begin{aligned}
        p(\bm{o}) = & \; \mathcal{N}(v_{\text{ego}}\vert \mu_v, \sigma_v) \, \mathcal{N}(a_{\text{long},\text{ego}}\vert 0, \sigma_a)  \\ & \; \mathcal{N}(\omega_{\text{ego}}\vert 0, \sigma_\omega) \, p_{\text{lat}}(y_{\text{ego}}) \, p_{\text{coll}}(\bm{o})\, p_{\text{safe}}(\bm{o}) \,.
    \end{aligned}
\end{equation}

For $p_{\text{safe}}(\bm{o})$, the model specifically seeks to avoid states where, under the assumption that the other vehicle begins braking suddenly with a deceleration of $a_{\text{OV}, \min}$, the ego vehicle would fail to avoid a collision despite initiating maximum braking after a response time of $\SI{1}{s}$.

\subsection*{Surprise-based re-planning}

% Incremental planning
Our model assumes that on every time step, the agent  performs planning incrementally, unless it observes a surprising event, in which case it re-plans the full policy.
Practically, on every time step the agent takes the policy chosen in the previous time step, disregards its first action (as it was already executed) and assumes that the remaining actions constitute all but the last actions of the new policy. The above policy selection mechanism is then used to generate the last action of the new policy. 

% Optional re-planning
In parallel with incremental policy updates, on every time step the agent accumulates evidence in favor of a full re-plan, using the accumulation rate $\lambda$:
\begin{equation}\label{eq:Evidence_main}
    E_t = E_{t-1} + \lambda \epsilon_t \, .
\end{equation} 
Traditionally in evidence accumulation literature, the accumulation rate is referred to as drift rate, which is interpreted as the quality of the incoming evidence~\cite{ratcliff2016diffusion} and can be linked to the decision-maker's efficiency of processing the perceptual information~\cite{metin2013adhd,reinhartz2023mechanisms}.
Here, we follow Engström et al.~\cite{engstrom_modeling_2024} in accumulating surprise $\epsilon \geq 0$, defined as the \emph{residual information} of the pragmatic value~\cite{dinparastdjadid_measuring_2023}:
\begin{equation}\label{eq:surprise}
    \epsilon_t = H \underset{\bm{o}}{\max} \ln p(\bm{o}) - \sum\limits_{\tau = t + 1}^{t+H} g_{\text{pragm}}\left(\widetilde{q}_o(\bm{o}_{\tau}\vert \bm{\pi}_t, q(\bm{s}_t))\right) \, .
\end{equation}
Under this definition, the surprise is the difference between the highest pragmatic value possible and the actual pragmatic value of a policy. For instance, if a policy would result in the most preferred observation, the agent would not accumulate any additional evidence. However, if a policy would lead to an undesired (i.e., \emph{a priori} unlikely) observation, much evidence for a re-plan would be gained.

If the accumulated evidence is below the threshold of $1$, the model then follows the extended policy. Otherwise, it generates a completely new policy, applying the above policy selection mechanism to every action in the policy. 

% World update
Finally, after the policy is determined, its first action $\bm{a}_t$ is used to update the state of the actual world. This is done using the state transition function of the \emph{generative process}:
\begin{equation}
    \bm{\eta}_{t+1} \sim \widehat{p}(\bm{\eta}' \vert \bm{\eta}_{t}, \bm{a}_{t})
\end{equation}

\subsection*{Metrics}
\subsubsection*{Front-to-rear scenario}
In this scenario, we used the vehicle trajectory and driver input data to extract brake response times and deceleration magnitudes. To that end, we followed Markkula et al.~\cite{markkula_farewell_2016}, extracting response times by fitting a piecewise-linear function to the recorded velocity data; the brake response time was then determined as the time instant when the first constant line switches to one with falling velocity. The slope of this line was used as the estimated deceleration magnitude.

When evaluating the goodness of fit of a model to the ground-truth human data, we used the mean absolute error $I$. 
Given a dataset $D_0 = \{(x, y) \mid x\in X_0, y\in Y_0\}$ and a linear approximation of the ground truth $y_{\text{data}}(x) = a_{\text{data}} x + b_{\text{data}}$ over the support interval $[x_0, x_1]$, we first discarded all sample pairs from $D_0$ for which $x \notin [x_0, x_1]$, resulting in $D$, after which we computed the residuals 
\begin{equation}
E = \{y - y_{\text{data}}(x) \mid (x, y) \in D \}\,.
\end{equation}
We assumed the residuals can be approximated by a linear function
\begin{equation}
e(x) = \alpha x + \beta + \varepsilon, \quad \varepsilon \sim \mathcal{N}(0, \sigma_E)\,,
\end{equation}
where $\sigma_E$ is the empirical standard deviation of $E$. 
We placed Gaussian priors
\begin{equation}
p_0(\alpha) = \mathcal{N}\!\left(0, \tfrac{\sigma_E}{2\sigma_X}\right), \quad 
p_0(\beta) = \mathcal{N}\!\left(0, \tfrac{\sigma_E}{2}\right)\,,
\end{equation}
with $\sigma_X$ denoting the empirical standard deviation of $X$. 
The posterior $p(\alpha, \beta \mid D)$ was then obtained using standard Bayesian linear regression~\cite{box2011bayesian}. 
Defining
\begin{equation}
I(\alpha, \beta) = \frac{1}{x_1-x_0} \int_{x_0}^{x_1} \left| \alpha x + \beta \right| \, \mathrm{d}x,
\end{equation}
we estimated the posterior mean $\mu_I = \mathbb{E}_{p(\alpha,\beta\mid D)}[I(\alpha,\beta)]$ and standard deviation $\sigma_I = \sqrt{\mathbb{V}_{p(\alpha,\beta\mid D)}[I(\alpha,\beta)]}$ using Monte Carlo sampling from $p(\alpha,\beta \mid D)$ with $10000$ samples.

When analyzing response times (Figure~\ref{fig:Rearend}\textbf{d} and second row of Figure~\ref{fig:Ablation}), we set $x_0 = \SI{0.9}{s}$ and $x_1 = \SI{3.6}{s}$. Meanwhile, for the acceleration error (Figure~\ref{fig:Rearend}\textbf{e} and third row of Figure~\ref{fig:Ablation}), we set $x_0 = \SI{0}{s^{-1}}$ and $x_1 = \SI{1.0}{s^{-1}}$.

\subsubsection*{Opposite-direction lateral incursion scenario}
In this scenario, we extracted brake and steering response times following Johnson et al.~\cite{johnson2025looking}, by interpolating along the acceleration $a_{\text{long}}$ and steering angle $\delta$ data to find the time at which $\SI{-1}{ms^{-2}}$ and $\SI{0.0077}{rad}$ (i.e., a steering wheel angle of $5^{\circ}$) are exceeded, to extract brake and steering response times respectively.

When evaluating model fit of our model variants to the human data in this scenario, we used the Jensen-Shannon divergence (a measure of dissimilarity of two categorical distributions) to compare predicted outcomes and the Wasserstein distance (a proper metric) for reaction times. Given the relatively small size of the ground truth dataset, we applied nonparametric bootstrapping~\cite{tibshirani1993introduction} with 10000 resamples to estimate the sampling variability of each metric.

\subsubsection*{Intersection scenario}
To extract response times, here we followed the same approach as in the lateral incursion scenario. We only compared our model to the first collision instance shown to each participant in the experiment of Ziraldo et al.~\cite{ziraldo_driver_2020} to ensure that the underlying data represents natural reactions of human drivers. Importantly, the human steering reaction times were extracted based on steering wheel angles. However, as Ziraldo et al.~\cite{ziraldo_driver_2020} did not report how this relates to the steering angle of the vehicle (which our model works with), there might be some discrepancy here, as we had to rely on the relationship reported by Johnson et al.~\cite{johnson2025looking}. In this scenario we used the same bootstrapped metrics as in the lateral incursion scenario (i.e., Jensen-Shannon divergence for collision probabilities and Wasserstein distance for response times).

\subsubsection*{Statistic significance}

To determine whether one model variant provided a statistically significant better fit than another, we computed the signal-to-noise ratio of the difference between models, defined as the mean difference in metric values divided by its estimated standard deviation (using sampling from posterior distributions or bootstrapping as described above). A signal-to-noise ratio exceeding 3 was considered indicative of a meaningful difference, following standard statistical heuristics~\cite{casella2024statistical}. Here, the difference is computed as the mean metric difference between models, and under the assumption of independence, the standard deviation of the difference is approximated as the square root of the sum of the individual variances.

\subsection*{Model parameters}

\begin{table*}
    \centering
    \caption{An overview of the most essential parameters used in the full model, such as in the Cross Entropy Method (CEM) used for policy selection. For tuned parameters, we normally only tested three to four different parameters, with the drift rate $\lambda$ being the noticeable exception.}
    \begin{tabular}{|p{0.2\textwidth}|p{0.4\textwidth}|p{0.275\textwidth}|}
    \hline
    \rowcolor{black!66}
    \textcolor{white}{\textbf{Parameter}} & \textcolor{white}{\textbf{Description}} & \textcolor{white}{\textbf{Source}} \\ \hline
    \rowcolor{black!33}
    \multicolumn{3}{|l|}{State transition function $p(\bm{s}' \vert \bm{s}, \bm{a})$} \\ \hline
    $\Delta t = \SI{0.2}{s}$ & Time step size  & Model of routine driving~\cite{engstrom_resolving_2024} \\  \hline
    $\SI{4.2}{m} \times \SI{1.72}{m}$ & Vehicle size  & Matching~\cite{johnson2025looking} \\ \hline
    $w = \SI{3.65}{m}$ & Lane width  & Matching~\cite{johnson2025looking} \\ \hline
    $\sigma_{a,0} = \SI{3}{ms^{-2}}$ & Noise applied to other vehicle's acceleration during belief update & Tuned parameter \\ \hline
    $\sigma_{\omega,0} = \SI{0.4575}{ms^{-2}}$ & Noise applied to other vehicle's steering rate during belief update & Tuned parameter \\ \hline
    $\sigma_{a} = \SI{0.6}{ms^{-2}}$ & Noise applied to other vehicle's acceleration during behavior prediction & Tuned parameter \\ \hline
    $\sigma_{\omega} = \SI{0.0915}{ms^{-2}}$ & Noise applied to other vehicle's steering rate during behavior prediction  & Tuned parameter \\ \hline
    $H_{\text{n}} = 20$ & Prediction horizon for the \emph{projected normative probability}  & Tuned parameter \\ \hline
    \rowcolor{black!33}
    \multicolumn{3}{|l|}{Observation probability $p(\bm{o} \vert \bm{s})$} \\ \hline
    $\dot{\varphi}_0 = \SI{0.00215}{s^{-1}}$ & Looming threshold & Taken from~\cite{lamble_detection_1999} \\ \hline
    \rowcolor{black!33}
    \multicolumn{3}{|l|}{Preference function $p(\bm{o})$} \\ \hline
    $\mu_v = v_0$ & Mean of preference distribution of agent's velocity $v$  &   Model of routine driving~\cite{engstrom_resolving_2024} \\ \hline
    $\sigma_v = \SI{0.5}{ms^{-1}}$ & Standard deviation of preference distribution of agent's velocity $v$ & Tuned parameter \\ \hline
    $\sigma_a = \SI{0.1}{ms^{-2}}$ & Standard deviation of preference distribution of agent's acceleration $a$ & Tuned parameter \\ \hline
    $\sigma_{\omega} = \SI{0.02}{s^{-1}}$ & Standard deviation of preference distribution of agent's steering rate $\omega$   &  Tuned parameter \\ \hline
    $g_{LL} = - 15000$ & Pragmatic value for leaving the road &   Tuned parameter  \\ \hline
    $g_{LC} = - 1000$ & Pragmatic value for driving on a lane boundary or in opposing lanes & Tuned parameter  \\ \hline
    $g_{C} = - 10000$ & Pragmatic value after collision with relative velocity of $\SI{10}{ms^{-1}}$  & Tuned parameter  \\ \hline
    \rowcolor{black!33}
    \multicolumn{3}{|l|}{Policy selection} \\ \hline
    $H = 30$  & Prediction horizon  & Model of routine driving~\cite{engstrom_resolving_2024} \\ \hline
    $N = 75$  & Number of particles & Model of routine driving~\cite{engstrom_resolving_2024} \\ \hline
    $K = 10$ & Number of iterations of policy sampling & Model of routine driving~\cite{engstrom_resolving_2024} \\ \hline
    $M = 100$ & Number of sampled policies & Tuned parameter \\ \hline
    $\beta = 0.1$ & $\beta M$ samples with lowest EFE are used as base for next iteration & Model of routine driving~\cite{engstrom_resolving_2024} \\ \hline 
    $\mathcal{N}\left(0, \SI{5}{ms^{-2}}\right)$ & Distribution to sample accelerations $a_{\tau}$ in first CEM iteration & Model of routine driving~\cite{engstrom_resolving_2024} (mean) and Tuned parameter (std)  \\ \hline
    $\mathcal{N}\left(0, \SI{0.1}{s^{-1}} \right)$ & Distribution to sample steering rates $\omega_{\tau}$ in first CEM iteration & Model of routine driving~\cite{engstrom_resolving_2024} (mean) and Tuned parameter (std)  \\ \hline
    $a_0 = \SI{-0.1}{ms^{-2}}$ & Acceleration applied when no pedal is pressed & Logical consideration \\ \hline
    $\lambda = 10 ^ {-5.95} $ & Evidence accumulation drift rate & Tuned parameter \\ \hline
         
    \end{tabular}
    \label{tab:Parameters}
\end{table*}

\begin{figure*}
    \centering
    \includegraphics[]{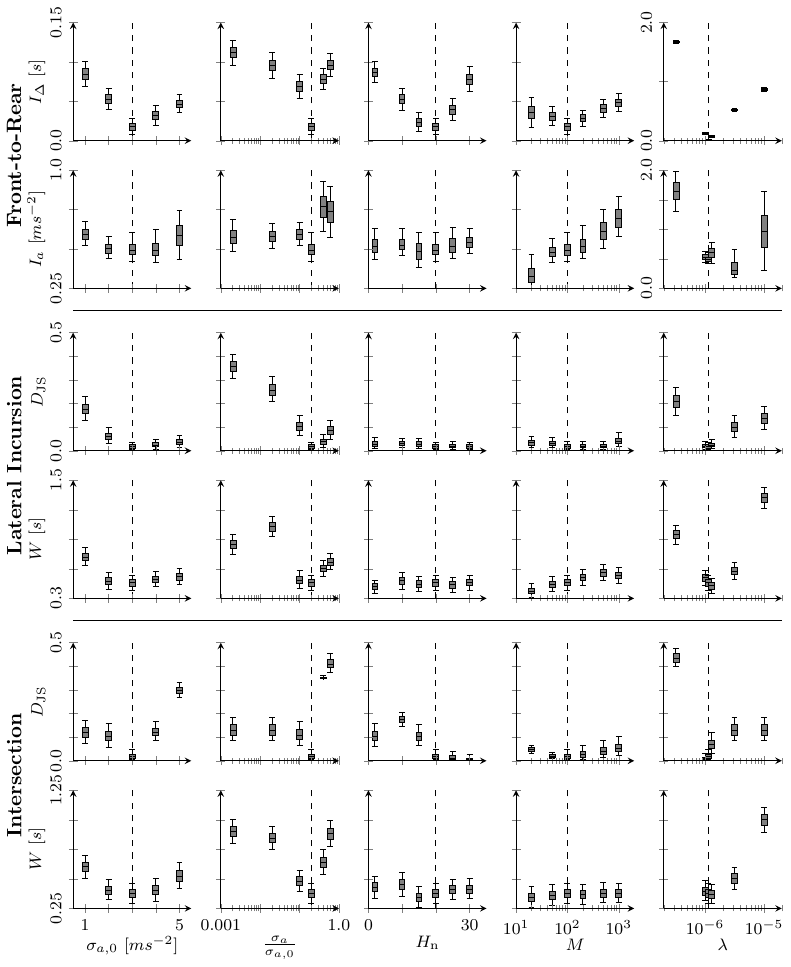}
    \caption{Sensitivity analysis for selected model parameters related to the state transition function and policy selection (Table~\ref{tab:Parameters}). The vertical dashed lines indicate the final values used in the full model. We generally used a scheme of only changing one parameter at a time. However, when changing $\sigma_{a,0}$ (middle column), we changed $\sigma_{\omega,0}$ with the same factor. Meanwhile, in the third column, we ensured that $\frac{\sigma_a}{\sigma_{a,0}} = \frac{\sigma_\omega}{\sigma_{\omega,0}}$, while keeping $\sigma_{a,0}$ and $\sigma_{\omega,0}$ fixed. In the box plots, the lower and upper whiskers indicate the 5th and 95th percentile, respectively. For $\lambda$ in the rightmost column, note the change in $y$-axis labels for some metrics.}
    \label{fig:SA_state}
\end{figure*}

\begin{figure*}
    \centering
    \includegraphics[]{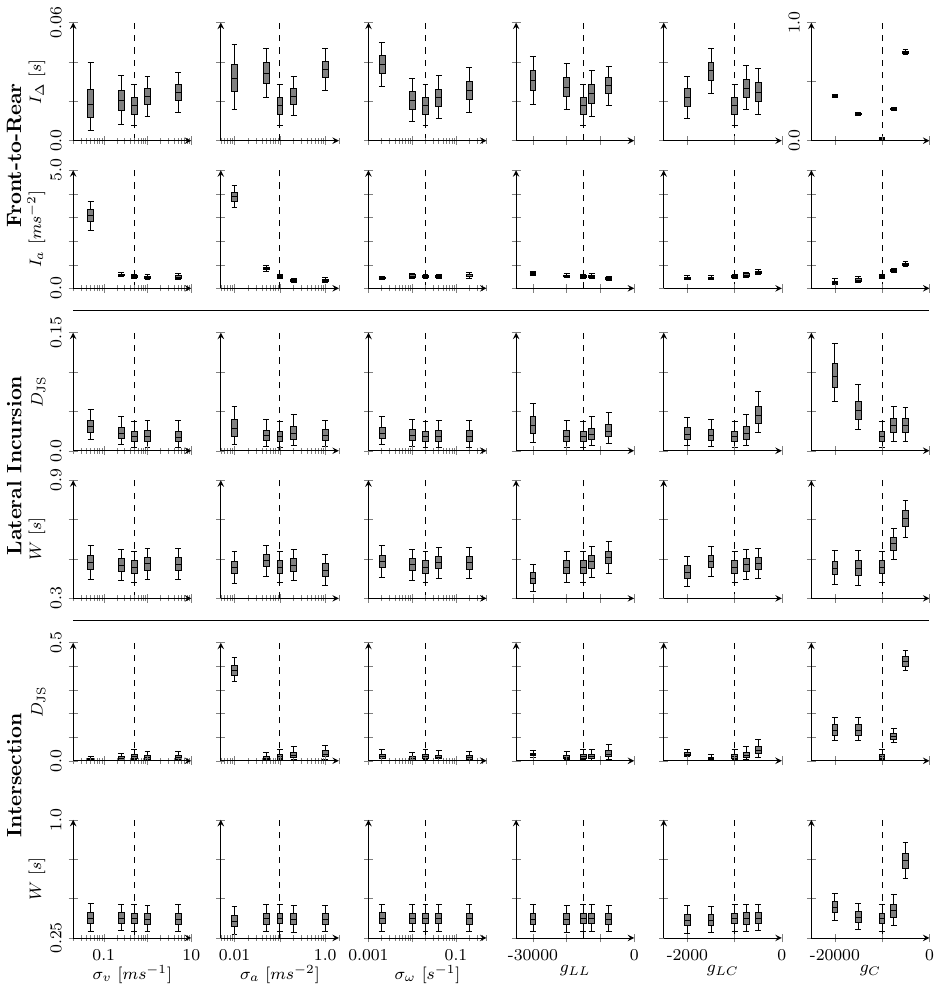}
    \caption{Sensitivity analysis for selected model parameters related to the preference function (Table~\ref{tab:Parameters}). The vertical dashed lines indicate the final values used in the full model. Similar to Figure~\ref{fig:SA_state}, we only changed one parameter at a time; the lower and upper whiskers in the box plots indicate the 5th and 95th percentile, respectively. While the metrics used are the same, the scales of the $y$-axes are different than in Figure~\ref{fig:SA_state} for illustrative purposes. For $g_C$ in the rightmost column, note the change in $y$-axis labels for $I_\Delta$.}
    \label{fig:SA_reward}
\end{figure*}

Each of the key model mechanisms has a number of parameters associated with it (see Table~\ref{tab:Parameters} for an overview of the 26 most essential parameters). For 12 of them, the exact values have been directly adopted from the literature, in particular the original model~\cite{engstrom_resolving_2024}. Another thirteen model parameters were treated as free parameters, and were manually tuned to qualitatively match the empirical observations. 

It was not the purpose of this study to obtain the closest possible fit to human data, so no exhaustive parameter optimization was performed. To analyze the impact that changes of model parameters have on its performance, we conducted a sensitivity analysis (Figures~\ref{fig:SA_state}~and~\ref{fig:SA_reward}). Specifically, we varied the thirteen tuned parameters from Table~\ref{tab:Parameters} one at a time, while keeping the other parameters fixed. For most parameters, this explored deviation by roughly one order of magnitude, if not otherwise constrained. For each of the new parameter sets, we reran all the simulation in the three scenarios described in the Results section, and calculated the goodness-of-fit metrics (Figure~\ref{fig:Ablation}). Using the methods described in the previous Metrics section to estimate the uncertainty of those metrics inherent in the limited number of simulations and human demonstrations, we express those in the corresponding box-plots.

The most consequential parameter is the evidence drift rate $\lambda$ (Equation~\eqref{eq:Evidence_main}), where moderate deviations can lead to large changes in reaction times and therefore substantially worse goodness of fit (the goodness of fit metrics $I_a$ and $W$ in Figure~\ref{fig:SA_state} can degrade on the order of seconds). Model behavior also deteriorates, as seen in the Jensen–Shannon divergences. This is expected, as for example a sufficient increase in $\lambda$ to trigger re-planning at (nearly) every timestep would in principle be equivalent to removing the evidence accumulation mechanism completely from the model, which was shown to be disadvantageous for reproducing human behavior in Figure~\ref{fig:Ablation}\textbf{b}. However, our chosen parameter value appears to involve a slight trade-off. In the lateral incursion and intersection scenarios, small increases in $\lambda$ can improve the accuracy of reaction times while worsening outcome prediction, whereas small decreases show the opposite trend. Meanwhile, the other hand-tuned policy selection parameter, the number of simultaneously evaluated policies $M$, had even weaker impact (Figure~\ref{fig:SA_state}). There, for $I_{\Delta}$ and $D_{\text{JS}}$, increasing or decreasing $M$ negatively affects the goodness of fit, but there was no evidence that these differences were statistically significant. Meanwhile, for $I_a$ and $W$, lower $M$ values did yield better fit to the data, but the improvements were marginal.

Parameters of the state transition function did have a major effect on model behavior (Figure~\ref{fig:SA_state}). For $\sigma_{a,0}$ and $\sigma_{\omega,0}$ (the noise assigned to the other vehicles' control actions in the generative model's state transition function), we generally observed a U-shaped response across all considered metrics, indicating that deviations from the chosen values led to poorer fit to the data, although not all differences were statistically significant. Similarly, $\sigma_{a}$ and $\sigma_{\omega}$ (the corresponding noise in the policy roll-out during behavior prediction, second column) chosen for the final model generally yielded best goodness-of-fit. The horizon for calculating the projected normative probability $H_{\mathrm{n}}$ was less influential: it only impacted model behavior in terms of front-to-rear reaction times ($I_{\Delta}$) and the prediction of collision outcomes in the intersection scenario ($D_{\mathrm{JS}}$). 

For the parameters of the preference function $p(\bm{o})$ (Equation~\eqref{eq:preference_likelihood}), two primary effects can be observed (Figure~\ref{fig:SA_reward}). First, the choice of $\sigma_v$ and $\sigma_a$ is important for producing a reasonable fit for the deceleration magnitude applied in the front-to-rear scenario ($I_a$ in Figure~\ref{fig:SA_reward}). Second, the choice of collision cost $g_C$ is critical, with slight deviations from the selected value leading to substantial decreases in model performance (Figure~\ref{fig:SA_reward}). Consequently, fine-tuning of this parameter is of high importance. Varying the remaining parameters did not result in statistically significant changes, suggesting that their coarse calibration is sufficient. However, this is somewhat expected, as the initial trajectories of the ego agents often are aligned with the preferences expressed in those parameters, which mostly influence the specific avoidance maneuver chosen, but not the specific reaction times. This can be seen in the specific metric expressing such behavior ($I_a$), where especially $\sigma_v$ and $\sigma_a$ have a significant impact.

Especially notable is that -- with parameters tuned only on the first two scenarios without any exposure to the intersection scenario -- the chosen parameters perform similarly well in the intersection scenario, suggesting good model generalizability. 
Taken together, this parameter sensitivity analysis indicates that 1) most of the sensitivity of the model is primarily concentrated in $\lambda$ and $g_C$; 2) most parameters are robust to minor perturbations; and 3) while not necessarily optimal, our final parameter settings achieve good fit to human data, with no parameter variations leading to substantially better goodness-of-fit across all three scenarios. 

\section*{Data availability statement}
The simulation data generated with our active inference model and its ablations in this study have been deposited in the OSF database (\url{https://osf.io/gs4bu}). As mentioned in the corresponding captions, videos of the results shown in Figures~\ref{fig:Rearend},~\ref{fig:Oncoming},~\ref{fig:Intersection} are provided in the Supplementary Materials. Source data underlying the figures is available as a Source Data file. 

\section*{Code availability statement}
The code used for the data analysis and modeling in this study is available at \url{https://github.com/tud-hri/Active-Inference-Collision-Avoidance} with \doi{10.5281/zenodo.20049511}. The code is provided under a non-commercial license (available at \url{https://github.com/tud-hri/Active-Inference-Collision-Avoidance/blob/main/LICENSE}) that permits use for research, teaching, personal experimentation, and scientific publication. This includes use of the licensed materials for benchmarking in academic or applied research publications.

\bibstyle{nature}
% \bibliography{zotero_reference, manual_bib}

\section*{Acknowledgments}
We thank Daphne Cornelisse for creating Figures~\ref{fig:Overview}~and~\ref{fig:architecture_sketch}, as well as Martijn Wisse, Todd Hester, Ran Wei, and Trent Victor for providing helpful feedback on the manuscript.

\section*{Author contributions statement}
\textbf{JFS}: Methodology, Software, Validation, Formal analysis, Investigation, Writing - Original Draft, Visualization. \textbf{JE}: Conceptualization, Validation, Writing - Original Draft, Supervision, Project administration, Funding acquisition. \textbf{LJ}: Formal analysis, Writing - Review \& Editing, Visualization. \textbf{MO'K}: Writing - Review \& Editing. \textbf{JM}: Writing - Review \& Editing. \textbf{JK}: Writing - Review \& Editing, Supervision. \textbf{AZ}: Resources, Writing - Review \& Editing, Supervision, Project administration, Funding acquisition.

\section*{Competing interests statement}
JE, MO'K, LJ and JM were employed by Waymo LLC and conducted the research without any external funding from third-parties. Techniques discussed in this paper may be described in U.S. Patent Application Nos. 18/614,428 and 63/657,623. Delft University of Technology received funding from Waymo LLC for parts of the research carried out by JFS, but JFS received no direct financial benefit for his contributions to the paper. The remaining authors declare no competing interests.

\end{document}